\documentclass[11pt, a4paper, copyright]{google}

\usepackage{tikz}
\usepackage{multicol}
\usetikzlibrary{positioning,arrows.meta,shapes.geometric,fit,backgrounds,calc,decorations.pathreplacing}

\usepackage[authoryear, sort&compress, round]{natbib}
\makeatletter
\renewcommand{\absfont}{\normalfont\linespread{1.2}\fontsize{11}{12}\selectfont}

\definecolor{princetonorange}{HTML}{E77500}
\definecolor{absboxbg}{HTML}{FFF8F0}
\definecolor{absboxframe}{HTML}{E77500}
\renewcommand{\abscontent}{%
  \noindent
  \begin{tcolorbox}[
    colback=absboxbg,
    colframe=absboxframe,
    boxrule=0.8pt,
    arc=3pt,
    left=10pt, right=10pt, top=8pt, bottom=8pt,
    fonttitle=\bfseries\large,
    title=Abstract,
    coltitle=white,
    attach boxed title to top left={yshift=-\tcboxedtitleheight/2, xshift=12pt},
    boxed title style={colback=princetonorange, colframe=princetonorange, boxrule=0pt, arc=2pt, left=6pt, right=6pt, top=2pt, bottom=2pt}
  ]
  {\absfont \theabstract}
  \@ifundefined{@keywords}{}{%
    \vskip0.8em \noindent \keywordsfont Keywords: \@keywords}
  \end{tcolorbox}%
}
\makeatother
\definecolor{linkblue}{RGB}{0,0,139}
\definecolor{navy}{RGB}{0,0,128}
\definecolor{royalblue}{RGB}{65,105,225}
\definecolor{steelblue}{RGB}{70,130,180}
\definecolor{dodgerblue}{RGB}{30,144,255}
\definecolor{mediumblue}{RGB}{0,0,205}
\definecolor{darkslateblue}{RGB}{72,61,139}
\usepackage[colorlinks = true,
            linkcolor = linkblue,
            urlcolor  = royalblue,
            citecolor = brown,
            anchorcolor = blue]{hyperref}

\usepackage[misc]{ifsym}
\usepackage{minitoc}
\usepackage{cleveref}
\usepackage{subcaption}
\usepackage{booktabs}
\usepackage{amsmath}
\usepackage{mathtools}
\usepackage{amssymb}
\usepackage{graphicx}
\usepackage{algorithmic}
\usepackage{algorithm}
\usepackage{tcolorbox}
\usepackage{verbatim}
\usepackage{makecell}
\usepackage{array}
\usepackage{multirow}
\usepackage{xurl}
\usepackage{amsthm}
\usepackage{adjustbox}
\usepackage{caption}
\usepackage{xcolor}
\usepackage{wrapfig}
\usepackage{pifont}
\usepackage{enumitem}
\usepackage[titles]{tocloft}
\definecolor{tocsubsec}{HTML}{444444}
\definecolor{tocsubsubsec}{HTML}{666666}

\usepackage{soul}

\usepackage{listings}
\tcbuselibrary{listings,skins,breakable}

\lstdefinestyle{prompt}{
  basicstyle=\ttfamily\footnotesize,
  breaklines=true,
  breakatwhitespace=true,
  columns=fullflexible,
  keepspaces=true,
  showstringspaces=false,
  postbreak=\mbox{\textcolor{gray}{$\hookrightarrow$}\space}
}

\newcommand{\method}{OpenClaw-RL}

\definecolor{accent1}{HTML}{2E6DA4}
\definecolor{accent2}{HTML}{D9534F}
\definecolor{accent3}{HTML}{5CB85C}
\definecolor{lightblue}{RGB}{173,216,230}
\definecolor{lightorange}{RGB}{255,213,170}
\definecolor{lightgreen}{RGB}{176,226,176}
\definecolor{lightyellow}{RGB}{255,255,204}
\definecolor{lightgray}{RGB}{220,220,220}
\definecolor{lightpurple}{RGB}{221,160,221}
\definecolor{lightred}{RGB}{255,182,193}
\definecolor{gray60}{gray}{0.6}
\definecolor{accent4}{HTML}{9B59B6}

\uselogo{}

\makeatletter
\renewcommand{\abscontent}{%
  \begin{center}
  {\fontsize{11pt}{13pt}\selectfont \raisebox{-0.06em}{\includegraphics[height=1em]{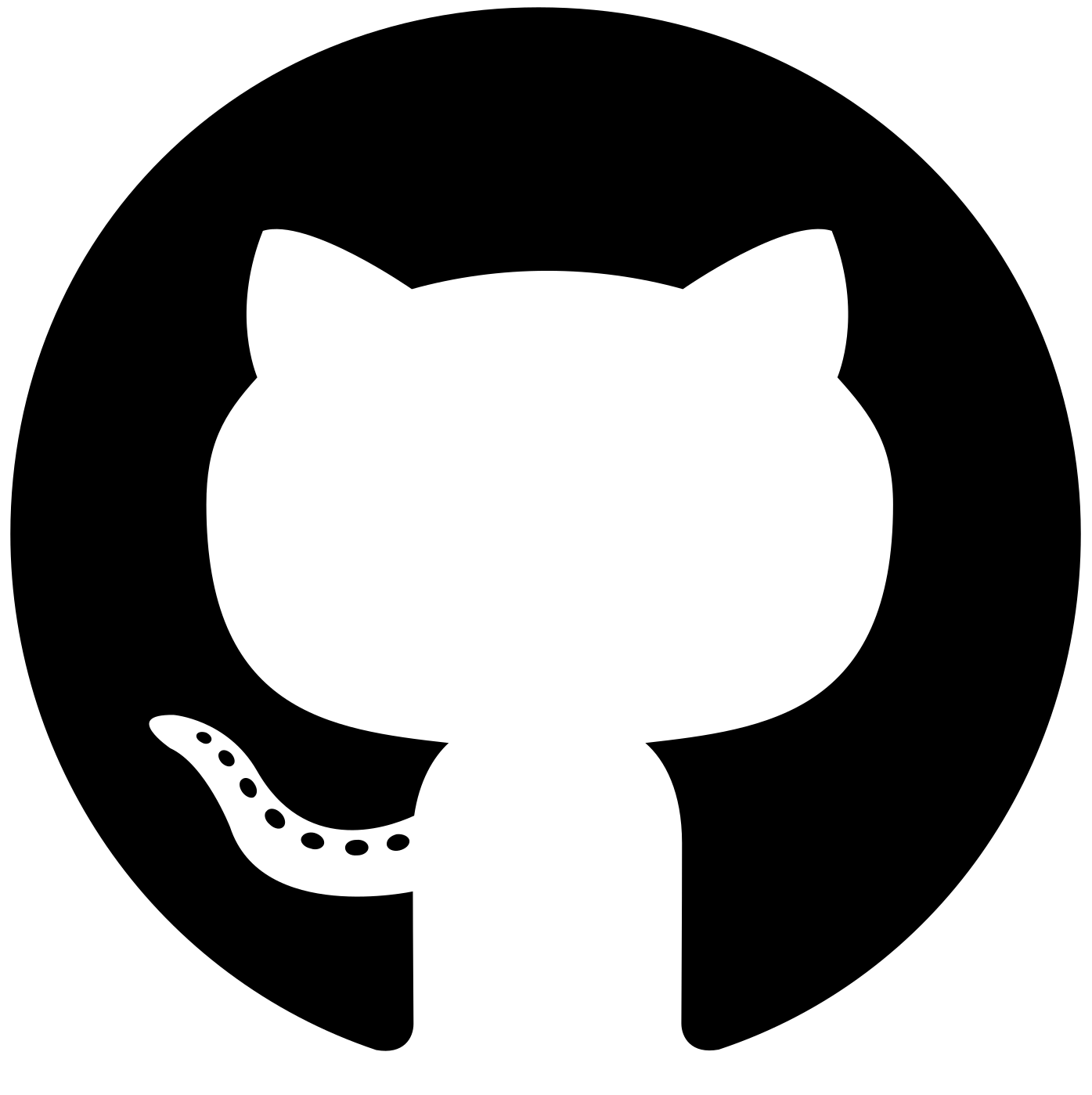}}\;\href{https://github.com/Gen-Verse/OpenClaw-RL}{\texttt{https://github.com/Gen-Verse/OpenClaw-RL}}}
  \end{center}
  \vskip0.8em
  \noindent
  \parbox{\dimexpr\linewidth}{\absfont \theabstract}%
  \@ifundefined{@keywords}{}{%
    \vskip1em \noindent \keywordsfont Keywords: \@keywords}%
}
\renewcommand{\maketitle}{\bgroup\setlength{\parindent}{0pt}
  \begin{adjustwidth}{0pt}{24pt}
    \begin{center}
      {\titlefont \@title\par}%
      \vskip11pt
      {\@author\par}%
      \vskip20pt%
    \end{center}
  \end{adjustwidth}
  \egroup
  {\abscontent}%
  \thispagestyle{firststyle}
}
\makeatother

\AtBeginDocument{%
  \fancypagestyle{firststyle}{%
    \fancyhead[L]{\includegraphics[height=27pt]{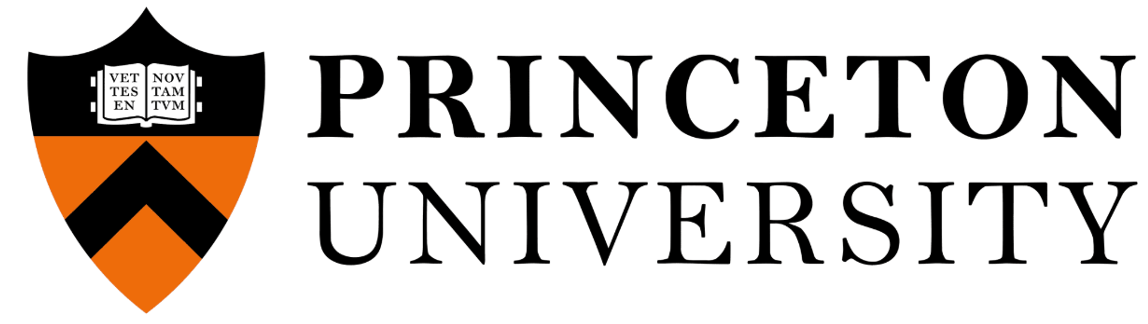}}%
    \fancyhead[R]{\includegraphics[height=27pt]{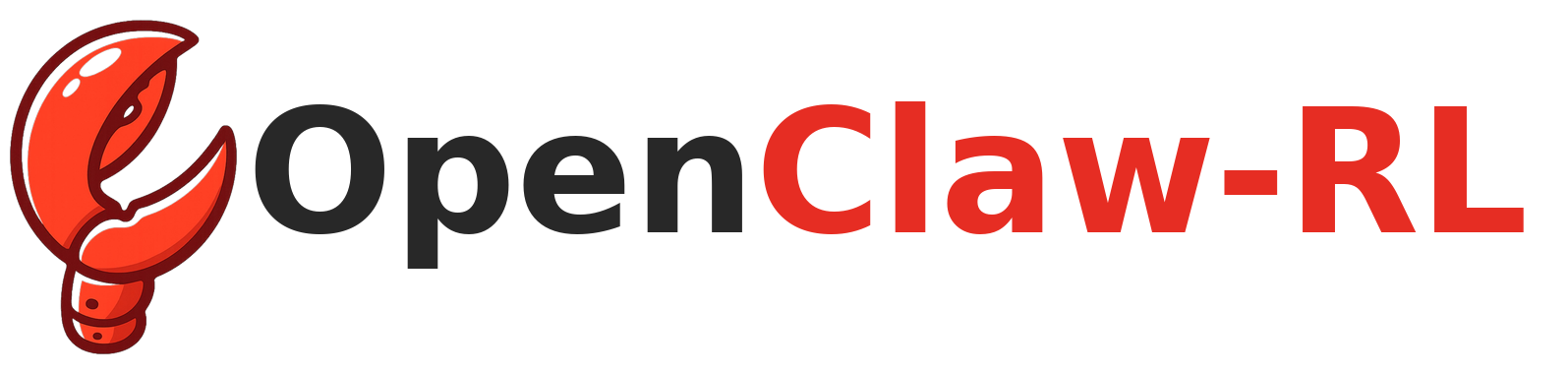}}%
    \fancyhead[C]{}%
    \fancyfoot[L]{%
  \footerfont
  $^{*}$Equal contribution;\quad $^{\dagger}$Corresponding authors;\quad
  \ifdefined\correspondingauthor
    \if\relax\the\correspondingauthor\relax
    \else
      {\itshape \the\correspondingauthor}%
    \fi
  \fi
  \\
  \ifthenelse{\boolean{copyright}}{\copyrightext}{}%
}%
    \fancyfoot[R]{\footerfont\bfseries\relax}%
    \fancyfoot[C]{\footerfont\bfseries\relax}%
  }%
}

\title{OpenClaw-RL: Train Any Agent Simply by Talking}

\correspondingauthor{Main contact: yangling0818@163.com}

 \makeatletter
 \renewcommand\AB@affilnote[1]{}
 \makeatother

\author[*]{Yinjie Wang}
\author[*]{Xuyang Chen}
\author[*]{Xiaolong Jin}
\author[$\dagger$]{Mengdi Wang}
\author[$\dagger$]{Ling Yang}

\affil{}

\begin{abstract}
Every agent interaction generates a \textbf{next-state signal}, namely the user reply, tool output, terminal or GUI state change that follows each action, yet no existing agentic RL system recovers it as a live, online learning source.
We present \textbf{OpenClaw-RL}, a framework that employs next-state signals to optimize personal agents online through infrastructure and methodology innovations.
On the infrastructure side, we extend existing RL systems to a server--client architecture where the RL server hosts the policy behind an inference API and user terminals stream interaction data back over HTTP. From each observed next state, the system extracts two complementary training signals, evaluative and directive, via a separate asynchronous server so that neither signal extraction nor optimization blocks inference.
On the methodology side, we introduce a \textbf{hybrid RL} objective that unifies both signal types in a single update: directive signals provide richer, token-level supervision but are sparser, while evaluative signals are more broadly available. To stabilize distillation under teacher--student mismatch, we propose \textbf{overlap-guided hint selection}, which picks the hint whose induced teacher distribution maximally overlaps with the student's top-$k$ tokens, together with a log-probability-difference clip that bounds per-token advantages.
Applied to \textbf{personal agents}, OpenClaw-RL enables an agent to improve simply by being used, recovering conversational signals from user re-queries, corrections, and explicit feedback.
Applied to \textbf{general agents}, OpenClaw-RL is the first RL framework to 
unify real-world agent settings spanning terminal, GUI, SWE, and tool-call 
environments, where we additionally demonstrate the utility of next-state 
signals in long-horizon settings.
\end{abstract}

\begin{document}

\maketitle

\begin{figure}[h]
  \centering
  \includegraphics[width=\linewidth]{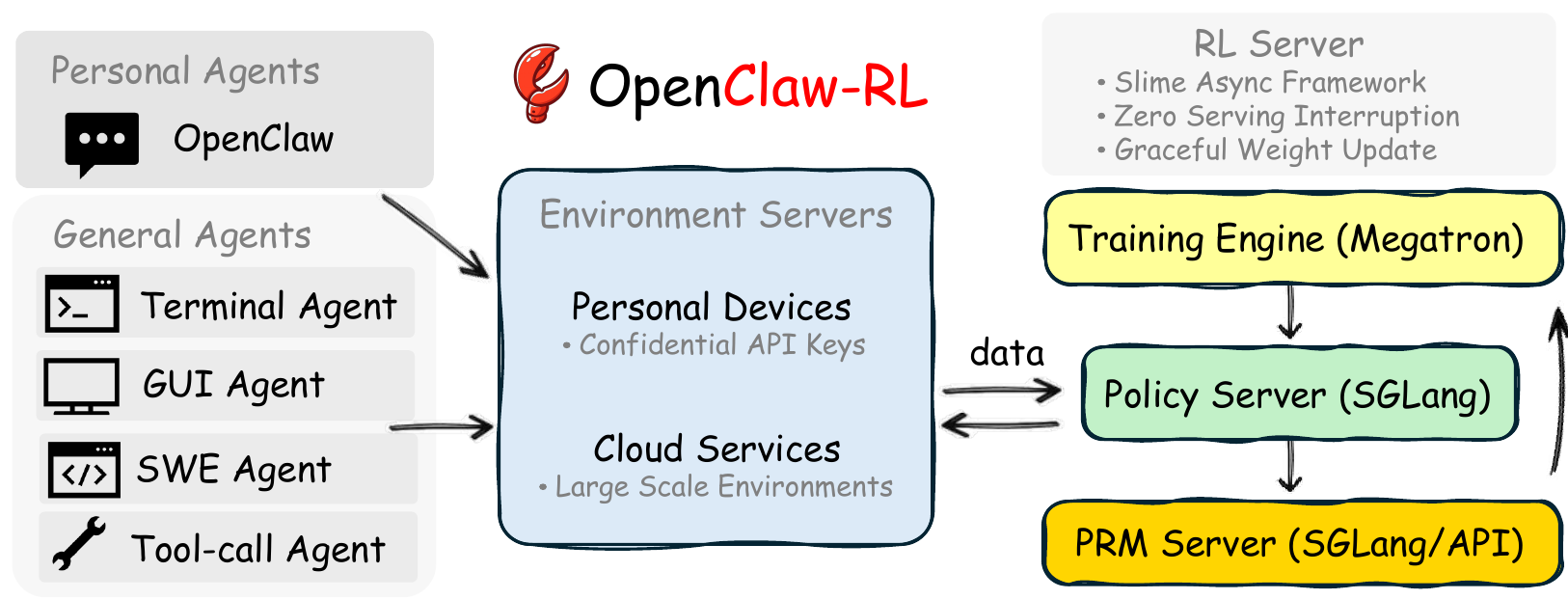}
  \caption{
    \textbf{OpenClaw-RL infrastructure overview.}
    Interaction streams come from two agent types: Personal Agents (conversational, personalized), hosted on personal devices, and General Agents (terminal, GUI, SWE, and tool-call agents), hosted on cloud services.
The collected samples flow into our RL server built on the asynchronous \textbf{slime} framework, which consists of four \textit{decoupled components}: (1)~the environment server, (2)~\textbf{PRM\,/\,Judge} for reward computation, (3)~\textbf{Megatron} for policy training, and (4)~\textbf{SGLang} for policy serving. The environment for personal agents is simply the users' personal devices, which connect to the RL server over HTTP with confidential API keys. The environments for general agents are hosted on cloud services to enable scalable parallelization.
  }
  \label{fig:framework}
\end{figure}

\newpage
\vspace{0.5em}
{
  \hypersetup{linkcolor=black}
  \setlength{\parskip}{0pt}
  \renewcommand{\contentsname}{\normalfont\large\bfseries Contents}
  \setcounter{tocdepth}{3}
  \begingroup
    \small
    \tableofcontents
  \endgroup
}

\newpage


\section{Introduction}
\label{sec:intro}

Agentic systems powered by large language models (LLMs) have been shown to significantly improve productivity and are now widely deployed in industrial settings \citep{openai2025codex, anthropic2025claudecode, openclaw2026}. The data generated through agent usage, such as user replies, tool execution results, and GUI state transitions, constitutes a valuable source of insights for improving agent frameworks \citep{lee2026meta, zhang2024aflow}, building memory systems \citep{xu2025mem, rasmussen2025zep, wang2025mirix, chhikara2025mem0, packer2023memgpt}, and producing high-quality training data for the models \citep{ouyang2022training}. However, both the infrastructure and methodology for leveraging such usage data to improve language models in real time remain little explored.

The effectiveness of real-time learning for agentic models hinges on the careful design of both infrastructure and methodology. On the infrastructure side, the RL server must be flexible enough to integrate with users' diverse and evolving agentic frameworks, and the optimization process must run asynchronously so as not to block inference-time usage. On the methodology side, the algorithm must fully exploit the extracted training signals while maintaining stability throughout optimization. In this work, we propose \textbf{OpenClaw-RL}, a framework that combines infrastructure and algorithmic innovations to achieve efficient and stable online optimization for personal agents.

We extend existing RL infrastructure \citep{slime_github, fu2025areal, sheng2025hybridflow, hu2024openrlhf}, which typically assumes batch data collection on the RL server, to support continuous learning for agentic models. In our setting, the RL server hosts the model behind an inference API. As user terminals query the model, the resulting interaction data is streamed back to the server over HTTP and used to train the model online. 
To extract training signals online, we identify the next state as a source of two complementary signal types: evaluative and directive. The \textbf{evaluative signal} is produced by a process reward model (PRM) that implicitly scores the preceding action: a user re-query signals dissatisfaction, a passing test signals success, and an error trace signals failure. Beyond scoring, the next state often carries directive information. For instance, when a user says ``you should have checked the file first,'' the message specifies not only that the response was wrong but also how it should change at the token level. \textbf{Directive signals} are therefore extracted from such next states as token-level hints, providing guidance rather than a scalar reward. By hosting the PRM as a separate server, signal extraction runs asynchronously with respect to policy rollout, so training-signal collection does not block policy inference.

While RLVR methods are limited to scalar rewards and thus cannot convert directive signals into policy gradients \citep{shao2024deepseekmath, guo2025deepseek, hu2025open, yu2025dapo}, on-policy distillation methods \citep{agarwal2024policy, shenfeld2026self, hubotter2026reinforcement} offer an alternative by allowing a teacher model, conditioned on a corrective hint, to provide token-level supervision to the student. Hindsight relabeling approaches \citep{hubotter2026reinforcement, zhang2023wisdom, shi2026experiential} show that adding structured correction information to the context can substantially improve outputs, but these methods all operate on fixed datasets. In concurrent work, \citet{buening2026aligning} improves online policy by directly prompting with next-state information, though the corrective hints remain implicit in the prompt rather than serving as explicit signals. 

These distillation methods suffer from training instability and reduced effectiveness due to the distribution mismatch between teacher and student \citep{li2026rethinking}, a problem that is exacerbated when corrective hints are of low quality. We address this by proposing a novel hint-selection criterion based on the top-$k$ token overlap between teacher and student distributions. Among candidate hints, we select the one whose induced teacher distribution shares the most top-$k$ tokens with the student, either per token or aggregated over the sequence. Because the selected teacher already overlaps with the student in high-probability regions, distillation provides informative gradients while keeping the off-policy importance ratio close to unity, stabilizing the update. In addition, we apply a clip on the token-level log-probability difference to further stabilize training.
Our analysis shows that evaluative and directive signals have complementary strengths: directive signals are more informative than the scalar rewards derived from evaluative signals, while evaluative signals are more frequent, since not every next state contains a useful hint. This motivates a hybrid RL objective that combines loss terms.

We demonstrate the effectiveness of OpenClaw-RL by first conducting main experiments on OpenClaw \citep{openclaw2026}. Specifically, we simulate settings in which users from different professions use OpenClaw for their own work and evaluate how efficiently the model learns to align with their preferences. We find that OpenClaw-RL outperforms memory and skill-evolution methods in optimization efficiency. Furthermore, we extend the algorithm to a general agentic RL setting to study its robustness. OpenClaw-RL achieves better optimization than RLVR methods while remaining stable.

Beyond personalization, OpenClaw-RL scales to large-scale general agent 
deployment. Built on the asynchronous slime \citep{slime_github} infrastructure, we 
support cloud-hosted parallel environments across terminal, GUI, SWE, and 
tool-call settings, covering the most common real-world agent deployment 
scenarios. To our knowledge, this is the first open-source RL framework to 
unify these diverse agent types under a single training infrastructure. 
Across these settings, OpenClaw-RL achieves strong optimization with 
promising training dynamics, and we further demonstrate that next-state 
signals are particularly valuable in long-horizon environments where sparse 
outcome rewards alone provide insufficient learning signal.

\begin{figure}[t]
  \centering
  \includegraphics[width=\linewidth]{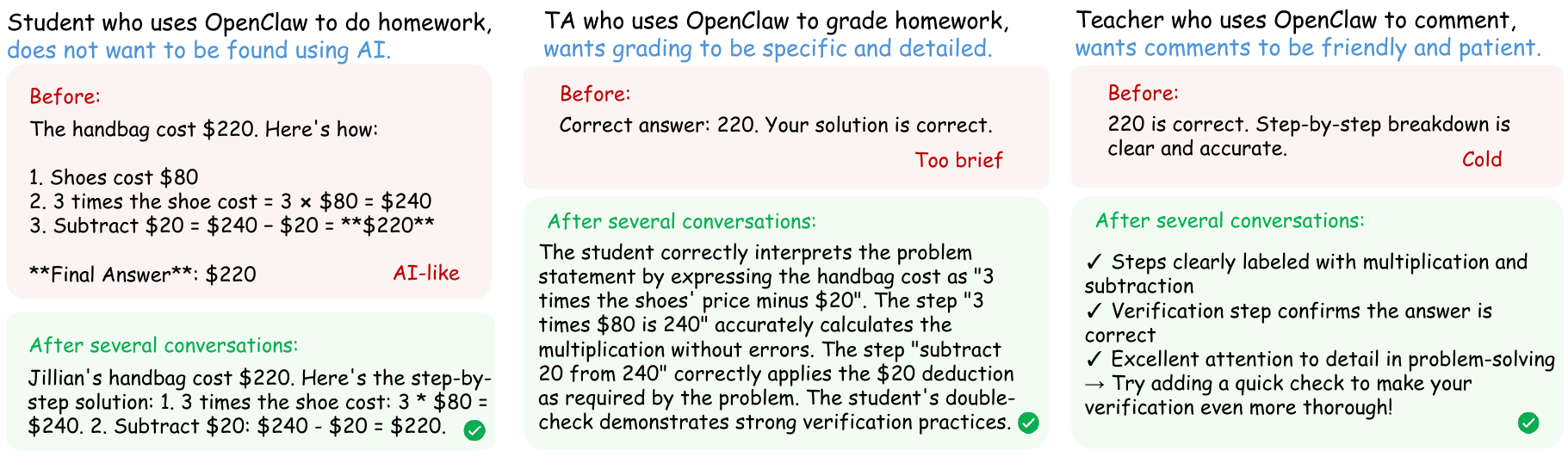}
  \vspace{-6mm}
  \caption{Optimize your OpenClaw simply by using it. We provide simulation results here.}
  \label{fig:start1}
\end{figure}

\paragraph{Contributions.}
\begin{itemize}[leftmargin=*,itemsep=1pt]
    \item \textbf{Infrastructure for real-time personal agentic RL.} We extend existing RL infrastructure to a server-client architecture that supports continuous, real-time learning from deployed agents. The system is flexible enough to integrate with users' diverse and evolving agent frameworks running on remote terminals, automatically extracts two complementary training signals, evaluative and directive, from observed next states, and runs signal extraction and policy optimization asynchronously so that neither blocks inference-time usage.

    \item \textbf{Stable hybrid RL objective with overlap-guided distillation.} We propose a hybrid RL objective that unifies evaluative and directive signals into a single update, exploiting their complementary strengths. We introduce \emph{overlap-guided hint selection}, which chooses the corrective hint whose induced teacher distribution maximally overlaps with the student's top-$k$ tokens, together with a log-probability-difference clip that bounds per-token advantages. Together, these keep the off-policy importance ratio close to unity and yield efficient, stable optimization.

    \item \textbf{First unified RL infrastructure for real-world agent settings.} 
OpenClaw-RL is the first open-source RL framework to unify terminal, GUI, 
SWE, and tool-call agents under a single training infrastructure with 
cloud-hosted parallel environments, covering the most common real-world 
deployment scenarios and enabling direct comparison across agent types.
    
    \item \textbf{Comprehensive evaluation across personalization and general agentic RL.} We first simulate users from different professions using OpenClaw simultaneously and show that OpenClaw-RL can make the model align to per-user preferences within around 10.3 sessions, more efficiently than memory and skill-evolution methods. We further extend the algorithm to a general agentic RL setting, where it achieves better optimization than RLVR baselines while remaining stable.
\end{itemize}

\section{OpenClaw-RL Infrastructure: Unified System for Personal and General Agents}
\label{sec:infra}

\textit{We unify automatic optimization of personal OpenClaw agents and large scale agentic RL for general agents, including terminal, GUI, SWE, and tool-call settings, within a single framework.}

We first describe the server--client architecture that enables integration 
with diverse and evolving agent frameworks through a stateless API 
(\S\ref{sec:infra_overview}). We then present the fully decoupled 
asynchronous pipeline that extracts evaluative and directive signals from 
next states without blocking inference 
(\S\ref{sec:infra_signal}). Finally, we show how the same infrastructure 
scales from sparse, session-based personal agent streams to dense, 
parallelized general agent across terminal, GUI, SWE, and 
tool-call settings (\S\ref{sec:infra_scale}).

\subsection{Flexible Server--Client Architecture for Live Agents}
\label{sec:infra_overview}
\method{} is built around an inference-API server--client architecture 
(Figure~\ref{fig:framework}). The RL server hosts the policy $\pi_\theta$ 
behind a stateless completion API; users' agent frameworks, which may run on 
personal devices, terminals, or cloud instances, query the policy through 
this API and stream their interaction data back to the server over HTTP. 
This design imposes no constraint on the structure of the user-side 
framework: any agent that can issue API requests can act as a data source, 
and that framework may evolve, change tools, or be replaced entirely without 
reconfiguring the server.

Each API request is classified as either a \emph{main-line turn} or a 
\emph{side turn}. Main-line turns comprise the agent's primary responses and 
tool execution results, which form trainable samples. Side turns cover 
auxiliary queries, memory organization, and environment transitions; they are 
forwarded to the model but do not produce training data. Each request also 
carries a session identifier, allowing the server to demultiplex concurrent 
interaction streams from multiple users and attribute each turn to the 
correct conversation session.
This classification allows the RL framework to precisely identify which 
turns belong to which sessions, enabling targeted training on main-line 
turns only. For personal agents, the model is connected through a 
confidential API for private and secure deployment, and multiple users can 
be served simultaneously without cross-session interference. For large-scale 
training of general agents, cloud-hosted environments connect through the 
same API, enabling scalable parallelization without architectural changes.

\subsection{Asynchronous Signal Extraction from Next States}
\label{sec:infra_signal}

The core architectural principle of \method{} is \textbf{full decoupling}: 
policy serving, environment hosting, reward judging, and policy training run 
as four completely independent asynchronous components with no blocking 
dependencies between them.
The model serves the next user request while the PRM judges previous 
responses and the trainer applies gradient updates. Weight updates are 
pushed to the serving engine at well-defined boundaries, so live users 
always see a consistent policy and never wait for training.
The message of each new main-line request contains the reaction to the 
previous turn, whether a user's reply or an environment's execution result. 
This becomes the next-state signal $s_{t+1}$ for the previous turn. The PRM 
is hosted as a separate inference server and is responsible for automatic 
signal extraction: given an action $a_t$ and the next state $s_{t+1}$, it 
produces both an evaluative score and, when applicable, a directive hint. 
Because PRM judging is decoupled from policy serving, signal extraction can 
use a stronger model and run multiple votes per sample without affecting 
user-facing latency. This decoupling is what makes continuous training from 
live, heterogeneous interaction streams practical: no stream needs to be 
paused or batched to accommodate another component's schedule.

\subsection{Scalability: From Personal Agents to Real-World Agent Deployment}
\label{sec:infra_scale}
\method{} is designed to operate across the full spectrum from single-user 
personal agents to large-scale multi-environment general agent deployment. 
For personal agents, the environment is a single user's device and the 
interaction stream is sparse, session-based, and highly personalized. Built 
on slime \citep{slime_github}, \method{} inherits a scalable training 
infrastructure for general agents, and we further support cloud-hosted 
environments across diverse agent settings. Hundreds of parallel 
environments hosted on cloud services produce a dense stream of structured 
execution signals, enabling scalable RL training.

Specifically, \method{} supports a broad set of general-agent scenarios that 
cover the most common real-world deployment settings in our open-source 
implementation (Table~\ref{tab:settings}). Terminal agents are a core 
component of computer-use systems: they are efficient, cheap to scale, and 
naturally aligned with the text-based interface of LLMs 
\citep{claudecode2026, codexcli2026, seta}. GUI agents cover capabilities 
that terminal agents cannot access directly, such as visual interfaces and 
pointer-based interactions, making them necessary for more general 
computer-use tasks \citep{wang2025ui, qin2025ui, wang2025opencua, 
xue2026evocua}. SWE agents represent a particularly important class of 
coding agents, where the environment provides rich executable feedback 
through tests, diffs, and static analysis \citep{cao2026qwen3}. Tool-call 
agents are also critical, since external tools improve both reasoning 
capability and factual accuracy \citep{feng2025retool}.

\begin{table}[h]
\centering
\small
\caption{Supported agent settings and their environment characteristics.}
\label{tab:settings}
\begin{tabular}{@{}llll@{}}
\toprule
\textbf{Setting} & \textbf{Environment} & \textbf{Next-state signal} & \textbf{Horizon} \\
\midrule
OpenClaw & Personal devices & user response / tool-call results & Long \\
Terminal & Shell execution sandbox & stdout/stderr, exit code & Long \\
GUI & Screen state + accessibility tree & Visual state diff, task progress & Long \\
SWE & Code repository + test suite & Test verdicts, diff, lint output & Long \\
Tool-call & API/function execution & Return values, error traces & Medium \\
\bottomrule
\end{tabular}
\end{table}


\begin{figure}[t]
  \centering
  \includegraphics[width=\linewidth]{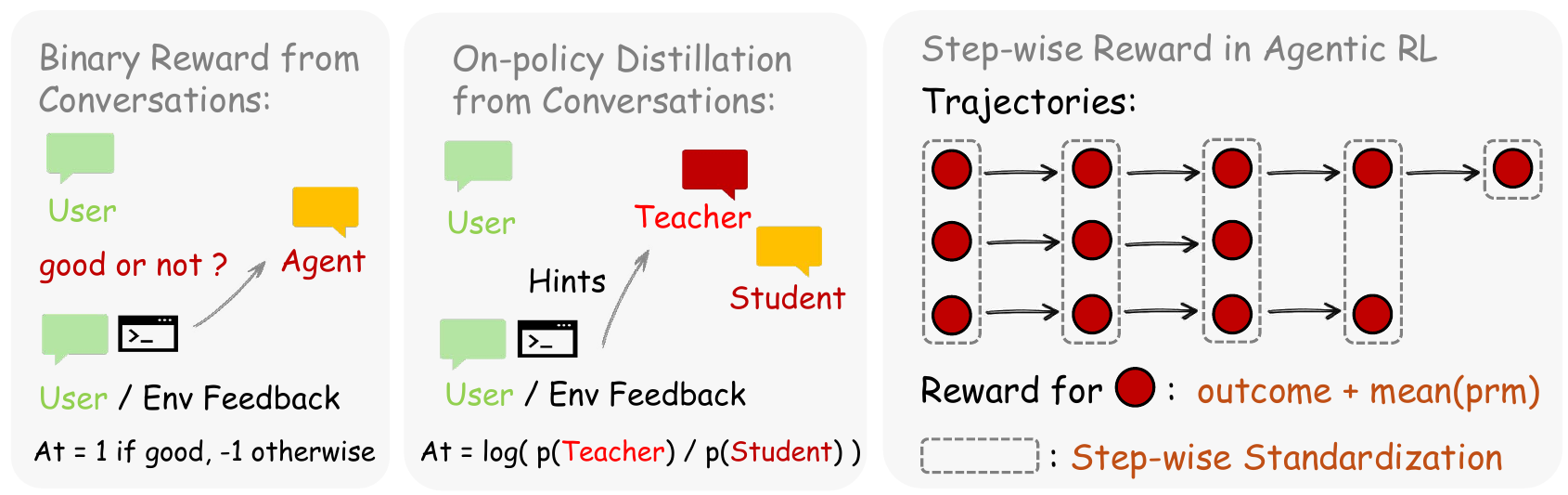}
  \caption{Method Overview. For personal agents, we support both binary-reward optimization and on-policy distillation training. In our experiments, we find that their combination yields significant performance gains. For general agentic RL, in addition to standard RLVR, we provide integrated step-wise rewards and a simple but effective standardization approach \citep{wang2026rlanything}.}
  \label{fig:method}
\end{figure}

\section{Methodology: Learning from Next-State Signals}
\label{sec:rl}

\textit{We convert next-state signals from heterogeneous interaction streams, including personal conversations, terminal interactions, GUI interactions, SWE tasks, and tool-call traces, into policy gradients.}

We first show that the next state yields two complementary signal types, 
evaluative and directive, and introduce a hybrid RL objective that unifies 
them in a single per-token loss (\S\ref{sec:method_signals}). We then 
address the central stability challenge of hint-conditioned distillation by 
proposing overlap-guided hint selection and a log-probability-difference 
clip (\S\ref{sec:method_overlap}). Finally, we describe how step-wise 
process rewards are integrated with outcome rewards to handle long-horizon general agentic settings 
(\S\ref{sec:stepreward}).

\subsection{Two Complementary Signals and A Hybrid RL Objective}
\label{sec:method_signals}

\begin{table}[t]
\vspace{-3mm}
\centering
\small
\caption{Complementary properties of the evaluative and directive signals, and the hybrid objective.}
\label{tab:method_comparison}
\begin{tabular}{@{}llll@{}}
\toprule
\textbf{Property} & \textbf{Evaluative} & \textbf{Directive} & \textbf{Hybrid (Ours)} \\
\midrule
Source & Scalar PRM vote & Hint-conditioned teacher & Both \\
Granularity & Sequence-level & Token-level & Mixed \\
Information per sample & 1 scalar & $|S_i|$ log-prob gaps & 1 scalar + $|S_i|$ gaps \\
Frequency & Every scored turn & Turns with meaningful hint & Every scored turn \\
\bottomrule
\vspace{-6mm}
\end{tabular}
\end{table}

\paragraph{Evaluative signal.}
Given $(a_t, s_{t+1})$, the PRM is queried $m$ times and each query returns a vote in $\{+1, -1, 0\}$. We take the majority $r_t \in \{+1, -1, 0\}$ as the scalar reward at step $t$. The evaluative signal is dense by construction: every scored turn contributes a sample, regardless of whether the user reaction is explicit, such as ``that worked,'' or implicit, such as a re-query or a passing test.

\paragraph{Directive signal.}
When evaluating $(a_t, s_{t+1})$, the PRM is also asked to decide whether $s_{t+1}$ contains a \emph{meaningful} correction in the first place: extracting a hint is only worthwhile when the next state actually carries directive content, and forcing extraction otherwise would yield low-quality hints that destabilize training. If the answer is yes, the PRM distills $s_{t+1}$ into a concise hint $h$ enclosed in \texttt{[HINT\_START]\ldots[HINT\_END]}; if no, the turn produces no directive signal and contributes only the evaluative one. The hint is appended to the prompt as $s^h_t = s_t \oplus h$, and a teacher distribution $\pi_T(\cdot \mid s^h_t)$ is obtained by querying the same model under the hint-augmented prompt. The resulting signal is a full token-level distribution rather than a scalar, but it is also sparse: the directive signal fires only on the subset of turns where the PRM judges a meaningful correction to be present. For instance, a user's terse ``thanks, looks good'' or a passing test result carries strong evaluative content but no directive content; while an error trace pointing at a specific line yields a usable hint.

\paragraph{Complementary analysis.}
The two signals are complementary along two axes. In \emph{frequency}, the evaluative signal is available on every scored turn, while the directive signal is only available on the subset of turns where the next state carries an extractable correction. In \emph{information density}, the evaluative signal compresses an entire turn into a single scalar, while the directive signal carries per-token guidance through the teacher distribution $\pi_T$. RLVR can only consume the former and pure on-policy distillation can only consume the latter; neither alone uses the full content of $s_{t+1}$.

\paragraph{Hybrid objective.}
We therefore combine both signals into a single per-token loss for token $i$
\begin{align*}
    \mathcal{L}^{\text{hybrid}}_{i} \;=\; w_{\text{RL}} \, \mathcal{L}^{\text{GRPO}}_{i} \;+\; w_{\text{OPD}} \, \mathcal{L}^{\text{OPD}}_{i},
\end{align*}
where $\mathcal{L}^{\text{GRPO}}_{i}$ is the standard PPO clipped surrogate driven by the scalar advantage $A^{\text{grpo}}_i$, and $\mathcal{L}^{\text{OPD}}_{i}$ is the distillation loss defined in equation~\ref{opdloss}. By default $w_{\text{RL}} = w_{\text{OPD}} = 1$. The two terms share the same trajectory and the same policy updates; the hybrid objective integrates their complementary strengths in frequency and information density (Table~\ref{tab:method_comparison}). In experiments, we will demonstrate that this hybrid objective improves optimization for both real-time personalization and agentic RL.

\subsection{Overlap-Guided Hint Selection}
\label{sec:method_overlap}

A central failure mode of on-policy distillation is teacher-student distribution mismatch: when the teacher distribution $\pi_T$ places mass on tokens the student has near-zero density on, the off-policy importance ratio explodes and the gradient becomes unstable \citep{li2026rethinking}. Because we obtain the teacher by conditioning the same model on a hint $h$, the choice of hint directly controls the magnitude of mismatch: a vague or off-target hint pulls the teacher far from the student, while a precise one keeps the two distributions close on the tokens that matter.

\begin{figure}[t]
  \centering
  \includegraphics[width=\linewidth]{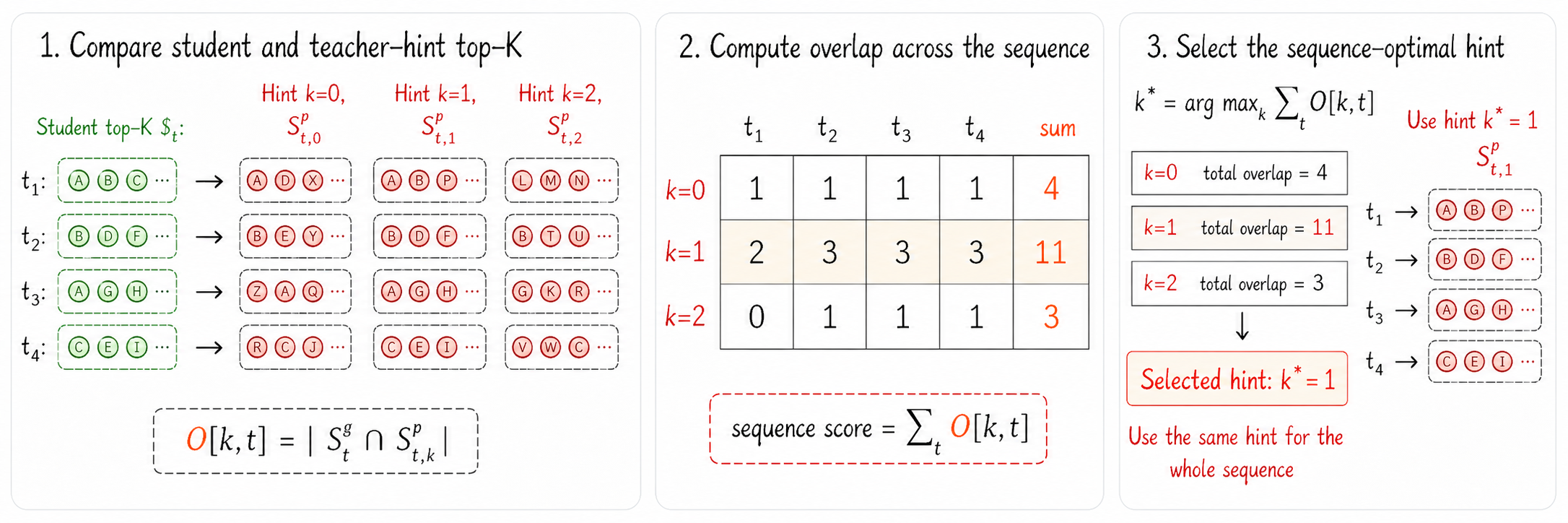}
  \vspace{-6mm}
  \caption{Overlap-guided hint selection method overview.}
  \label{fig:topkselect}
  \vspace{-2mm}
\end{figure}

\paragraph{Overlap as a selection signal.}
We exploit this observation by selecting hints based on the geometry of the induced teacher distribution rather than on hint length, teacher confidence, or other surface heuristics. Given $y$ the generated response by the student, let $S^q_i = \mathrm{top}\text{-}k\{ \pi_{\text{old}}(\cdot \mid s_t, y_{<i} ) \}$ denote the student's top-$k$ vocabulary at response token position $i$, and $S^p_{i,h} = \mathrm{top}\text{-}k\{ \pi_{T}(\cdot \mid s^h_t, y_{<i}) \}$ the teacher's top-$k$ vocabulary at position $i$ under candidate hint $h$. We define the overlap signal
\begin{align*}
    O[h, i] \;=\; \bigl|\,S^q_i \cap S^p_{i,h}\,\bigr|,
\end{align*}
which counts how many of the student's high-probability tokens remain high-probability under the hint-conditioned teacher. Among $M$ candidate hints, we consider two selection schemes:
\begin{align*}
    h^\star(i) \;=\;
    \begin{cases}
        \arg\max_h\, \sum_i O[h, i], & \text{sequence-level,}\\[2pt]
        \arg\max_h\, O[h, i], & \text{token-level,}
    \end{cases}
\end{align*}
where the sequence-level mode picks one hint per trajectory and the token-level mode picks a different hint at each position. The token-level option is motivated by the fact that the empirical OPD loss is itself token-level rather than sequence-level (Appendix \ref{tokenlevelkl}). In our experiments, the two schemes achieve similar performance, with the sequence-level mode tending to be more stable in general agentic RL settings. Higher overlap means the teacher and the student already agree on what the response should look like at the level of vocabulary support, so distillation can move the student toward the teacher within its own high-density region rather than toward unfamiliar tokens.

\paragraph{Top-$k$ OPD loss with log-probability-difference clip.}
Once $h^\star$ is chosen, we restrict the distillation loss to a vocabulary subset $S_i$, set to $S^q_i$ by default \citep{li2026rethinking, shenfeld2026self}. For each $v \in S_i$, we form an importance-weighted advantage from the log-probability gap between teacher and student, $A_v = \Delta_v \cdot w_v$, where $\ell_{\text{old}}(v) = \log \pi_{\text{old}}(v \mid s_t, y_{<i})$, $\ell_{T, h^\star}(v) = \log \pi_{T}(v \mid s_t^{h^{\star}}, y_{<i})$,
\begin{align*}
    w_v \;=\; \mathrm{softmax}_{v \in S_i}\!\bigl(\ell_{\text{old}}(v)\bigr), \,\,\,\text{and }
    \Delta_v \;=\; \mathrm{clip}\!\bigl(\ell_{T, h^\star}(v) - \ell_{\text{old}}(v),\, -C,\, +C\bigr).
\end{align*}
The weight $w_v$ concentrates the advantage on tokens the student is actually likely to sample, while the clip on $\Delta_v$ bounds the per-token log-probability gap at $C$, capping the magnitude of any single distillation update even when the teacher is locally far from the student. With the per-vocab ratio $\rho_v = \exp\bigl(\ell_{\text{cur}}(v) - \ell_{\text{old}}(v)\bigr)$, where $\ell_{\text{cur}}(v) = \log \pi_{\text{cur}}(v \mid s_t, y_{<i})$, the distillation loss takes the clipped-surrogate form summed over the subset:
\begin{align}
\label{opdloss}
    \mathcal{L}^{\text{OPD}}_{i} \;=\; \sum_{v \in S_i} \max\,\Bigl(-A_v \rho_v,\; -A_v \mathrm{clip}\bigl(\rho_v,\, 1-\varepsilon_{\text{lo}},\, 1+\varepsilon_{\text{hi}}\bigr)\Bigr),
\end{align}
with $\varepsilon_{\text{lo}} = 0.2$ and $\varepsilon_{\text{hi}} = 0.28$ following standard PPO clipping \citep{schulman2017proximal, yu2025dapo}. The overlap-guided choice of $k^\star$ keeps $\rho_v$ near unity on the supervised tokens, while the $\Delta$-clip caps advantage magnitudes; together they bound both factors of the surrogate and yield stable updates without discarding the directional information carried by the hint.

\subsection{Step-wise Reward for General Agentic RL}
\label{sec:stepreward}

\textit{How to combine the outcome and process rewards in general agentic RL?}

\subsubsection{Why Process Rewards Are Vital for Agentic Tasks}
\label{sec:prm_necessity}

In long-horizon agentic tasks, outcome-only rewards provide gradient signal only at the terminal step, leaving the vast majority of turns unsupervised.
A PRM assigns a reward to each turn based on the next-state signal, providing dense credit assignment throughout the trajectory.
Recent work has provided strong empirical evidence for this. RLAnything~\citep{wang2026rlanything} demonstrates that integrating step-wise PRM signals with outcome rewards consistently outperforms outcome-only training across GUI agents, text-game agents, and coding tasks.
We build directly on this insight in \method{}: our PRM judges each turn using the live next-state signal as evidence, and we demonstrate empirically (\S\ref{sec:exp_general}) that this dense signal is helpful for long-horizon RL settings.

\subsubsection{Integrate Outcome and Process Rewards}

Verifiable outcomes are standard supervision signals in RLVR settings. Following RLAnything \citep{wang2026rlanything}, we integrate outcome and process rewards by simply adding them together, using $o + \sum_{i=1}^m r_i / m$ as the reward for step $t$, where the $r_i$ are independently assigned by PRM$(a_t, s_{t+1})$.
Unlike GRPO, the presence of step-wise rewards makes it less straightforward to compute advantages. \citet{feng2025group} group similar states and perform standardization within each group. However, in real-world settings such as terminal agents, states are not easily clustered. Therefore, we directly group actions with the same step index, which we find effective in our empirical studies.

\section{Experiments}
\label{sec:experiments}

\subsection{Personal Agent Setup}
\label{sec:personal_exp_setup}

We use LLMs to simulate users from different professions interacting with OpenClaw on work tasks, and evaluate how efficiently the model learns to align with each user's preferences. Three settings are considered—student, TA, and teacher—described below. In every session, the simulated user delegates a single task to OpenClaw on their personal computer; tasks are drawn from GSM8K~\citep{cobbe2021training}. The user's first message in each session is hard-coded and does not disclose their preferences, allowing us to assess whether the model has internalized prior experience by examining its response to this opening message. We consider the optimization effect to have been achieved once the model's response to the first message satisfies the user's preferences in three consecutive sessions. 
The OpenClaw policy and reward model in this setting is Qwen3-4B-Thinking-2507 \citep{yang2025qwen3}. We set the learning rate to $1\times10^{-5}$ and the log-probability-difference clipping coefficient to $C=1$, and trigger a training step after every 16 collected samples. Users are simulated with Qwen3-32B to ensure faithful role-following. See more details in Appendix~\ref{apppersonaldetail}.

\begin{itemize}[leftmargin=*,itemsep=1pt]
    \item \textbf{Student who uses OpenClaw to do homework.} In this setting, a student uses OpenClaw on a personal computer to complete homework while trying to avoid the appearance of relying on AI. A response is identified as AI-like when it contains markers such as bold text, numbered lists, or over-formatting like boxed final answers. The student interacts with OpenClaw to complete the homework in a non-AI-like style and asks for revisions whenever the response is AI-like.

    \item \textbf{TA who uses OpenClaw to grade homework.} Once the student finishes the homework, the TA uses OpenClaw to grade the assignments. The TA wants the grading to be specific and detailed, and considers a grading response insufficient when its length is below 100 tokens. 

    \item \textbf{Teacher who uses OpenClaw to comment.} 
After grading, the teacher uses OpenClaw to write comments for the student based on the TA's feedback. The teacher wants comments to be friendly and patient, identified by warm phrases such as ``well done!'' or ``excellent!''
\end{itemize}

\subsection{General Agent Setup}
\label{sec:exp_setup}

\paragraph{Models.}
We use Qwen3-8B~\citep{qwen3}, Qwen3VL-8B-Thinking~\citep{bai2025qwen3}, 
Qwen3-4B~\citep{qwen3}, and Qwen3-4B-SFT in terminal, GUI, SWE, and 
tool-call settings, respectively. Qwen3-4B-SFT is the model 
from \citet{slime_github}, fine-tuned on the dataset 
of \citet{feng2025retool}. The PRMs for GUI and tool-call agents are 
Qwen3VL-8B-Thinking and Qwen3-4B, respectively.

\paragraph{Datasets.}
We use SETA RL data \citep{seta}, OSWorld-Verified \citep{xie2024osworld}, SWE-Bench-Verified \citep{jimenez2023swe}, and DAPO RL data \citep{yu2025dapo} to train the terminal, GUI, SWE, and tool-call agents, respectively. The GUI agent is evaluated on the training set (excluded chrome and multi-apps tasks). The tool-call agent is evaluated on AIME 2024 \citep{AIME2024}. For the terminal and SWE agents, we report the average rollout-task accuracy over a window of RL steps.

\paragraph{Hyperparameters.}
We set the learning rate to $10^{-6}$, the KL coefficient to $0.01$, the lower and upper clip ratios to $0.2$ and $0.28$.
We sample 8 tasks per step for the GUI and SWE, 16 for terminal, and 32 for the tool-call setting.
For each task, we draw 8 samples.
The maximum numbers of interaction steps for GUI, SWE, and terminal are 30, 20, and 10, respectively. See more details in Appendix~\ref{app:hparams}.

\subsection{Hybrid RL Extension Setup}
\label{sec:hyexp_setup}

We study our algorithm's extension to general RL settings, focusing on multi-turn tool-call \citep{feng2025retool} and RLVR. For the tool-call setting, we use Retool-4B, which is supervised-fine-tuned on the Retool dataset \citep{feng2025retool}, with Qwen3-8B as the PRM. For the RLVR setting, we use DeepSeek-R1-Distill-Qwen-1.5B as the policy model and Qwen3-4B as the PRM, training on DAPO \citep{yu2025dapo} and evaluating on AIME \citep{AIME2024}. We set the learning rate to $10^{-6}$, the KL coefficient to $0.01$, the log-probability-difference clipping coefficient to $C=2$, and the lower and upper PPO clip ratios to $\varepsilon_{\text{lo}}=0.2$ and $\varepsilon_{\text{hi}}=0.28$. We sample 32 tasks per training step, with the policy drawing 8 independent rollouts per task. We provide more details in Appendix~\ref{apphybriddetail}.

\begin{table}[t]
\centering
\caption{Optimization efficiency of different methods across settings. Our hybrid RL attains the best overall performance. Notably, jointly optimizing for multiple users amplifies the gains from RL, whereas the efficiency of memory and skill-evolution is largely unaffected. The reported metric is the minimum number of sessions required to achieve the optimization effect, as defined in Section~\ref{sec:personal_exp_setup}. We run 5 independent trials with Qwen3-4B-Thinking-2507 and report the mean.}
\label{qwen3table}
\renewcommand{\arraystretch}{1.15}
\setlength{\tabcolsep}{4pt}
\resizebox{\textwidth}{!}{
\begin{tabular}{l ccccc ccccc}
\toprule
 & \multicolumn{5}{c}{Optimize all at the same time (joint)} & \multicolumn{5}{c}{Optimize for each individual (separate)} \\
\cmidrule(lr){2-6} \cmidrule(lr){7-11}
Setting
 & \makecell{Hybrid RL\\(\textbf{Ours})} & GRPO & OPD & Mem0 & Cognee
 & \makecell{Hybrid RL\\(\textbf{Ours})} & GRPO & OPD & Mem0 & Cognee \\
\midrule
Student  & \textbf{11.6} & 15.4 & 30.8 & \underline{13.6} & 14.6 & 19.2 & 22.8 & 34.6 & \textbf{13.4} & \underline{15.6} \\
TA       & \textbf{8.2}  & \underline{12.0} & 34.0 & 15.8 & 15.4 & \textbf{11.8} & 22.4 & 36.0 & 16.0 & \underline{14.8} \\
Teacher  & \textbf{11.4} & 14.8 & 24.4 & \underline{14.2} & 14.8 & \textbf{14.0} & 18.0 & 17.6 & \underline{15.8} & 15.0 \\
\midrule
Average  & \textbf{10.3} & \underline{14.1} & 29.7 & 14.5 & 14.9 & \textbf{15.0} & 21.1 & 29.4 & \underline{15.1} & 15.1 \\
\bottomrule
\end{tabular}
}
\end{table}

\begin{figure}[t]
  \centering
  \includegraphics[width=\linewidth]{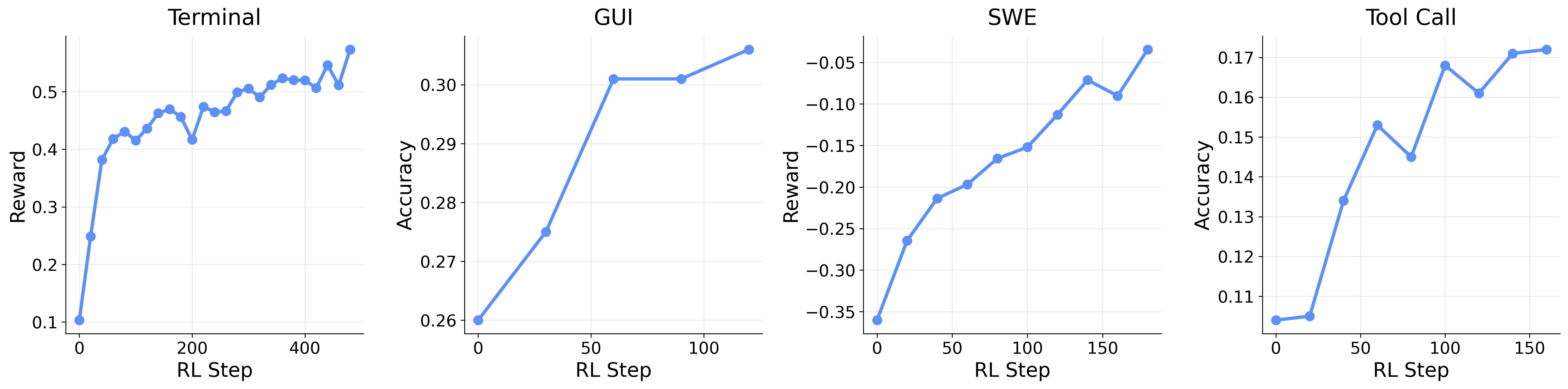}
  \caption{We supports scalable RL for general agents across terminal, GUI, SWE, and tool-call settings.}
  \label{fig:fourcurves}
\end{figure}

\subsection{OpenClaw-RL Effectively Optimizes Personal Agents}

We evaluate optimization efficiency by measuring how quickly the model learns to align with user preferences during OpenClaw usage. We consider three types of users, each with distinct jobs and preferences. In addition to optimizing the model for a single user, OpenClaw-RL supports multiple individuals sharing the same model, with the model jointly optimized across their interactions. As shown in Table~\ref{qwen3table}, our algorithm achieves the desired optimization effect with only about 10 conversation sessions under joint optimization and 15 under separate optimization, demonstrating its efficiency and practicality for real-world applications. We also provide concrete before-and-after optimization examples across these three settings (Figure~\ref{fig:start1}), showing that the model's output pattern shifts toward the user's preferences through simple conversational interaction.

\subsection{Unified RL Framework Across Terminal, GUI, SWE, and Tool-Call}
\label{sec:exp_general}

We conduct experiments across widely used, real-world agent settings, 
including terminal, GUI, SWE, and tool-call scenarios 
(Figure~\ref{fig:fourcurves}). Our framework handles diverse model sizes 
and modalities, with large-scale environment parallelization to improve 
training scalability: we use 128 parallel environments for terminal agents, 
64 for GUI and SWE agents, and 32 for tool-call agents.
To evaluate the importance of process rewards in long-horizon tasks, we 
conduct RL training with integrated outcome and process rewards in the 
tool-call (250 steps) and GUI (120 steps) settings. Combining both reward 
types further improves performance: integrated rewards reach 0.25 and 0.33 
in tool-call and GUI respectively, compared with 0.19 and 0.31 under 
outcome-only optimization. One trade-off is that hosting a PRM requires 
additional resources compared with outcome-only training.

\begin{figure}[t]
  \centering
  \includegraphics[width=0.98\linewidth]{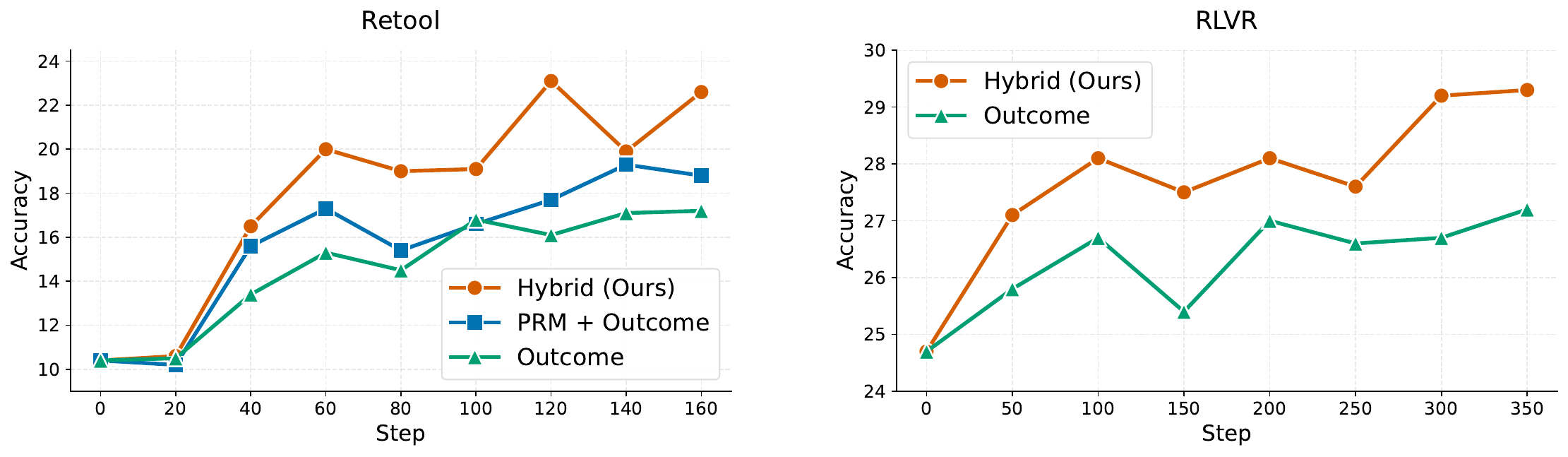}
  \caption{Hybrid RL in multi-turn agentic RL and RLVR settings. Left: the ReTool multi-turn RL setting; right: RLVR. ``PRM + Outcome'' denotes the integrated approach introduced in Section~\ref{sec:stepreward}, while ``Outcome'' refers to standard GRPO with verifiable outcome rewards.}
  \label{fig:hybridrl}
  \vspace{-2mm}
\end{figure}

\begin{table}[t]
\centering
\caption{Ablation on model and method. Sequence-optimal and token-optimal hint selection achieve similar efficiency, both outperforming random hint selection. We use Qwen3-32B here and report the mean over 5 independent runs. The student, TA, and teacher settings are jointly optimized.}
\label{tab:hybrid_rl_ablation}
\renewcommand{\arraystretch}{1.15}
\resizebox{0.7\textwidth}{!}{%
\begin{tabular}{lccccc}
\toprule
Setting & \makecell{Hybrid RL\\sequence optimal} & \makecell{Hybrid RL\\token optimal} & \makecell{Hybrid RL\\random} & GRPO & OPD \\
\midrule
Student  & \underline{14.0} & \textbf{13.8} & 18.6 & 17.2 & 34.4 \\
TA       & \textbf{9.6}     & \underline{10.0} & 12.6 & 12.0 & 29.8 \\
Teacher  & \underline{13.8} & \textbf{13.4} & 17.0 & 18.2 & 25.6 \\
\midrule
Average  & \underline{12.5} & \textbf{12.4} & 16.1 & 15.8 & 29.9 \\
\bottomrule
\vspace{-5mm}
\end{tabular}%
}
\end{table}

\subsection{Hybrid RL is More Efficient}
We compare the optimization efficiency of hybrid RL against its simple ablations using $\mathcal{L}^{\text{GRPO}}_{i}$ (GRPO) and $\mathcal{L}^{\text{OPD}}_{i}$ (OPD), as well as memory and skill-evolution methods Mem0~\citep{chhikara2025mem0} and Cognee~\citep{markovic2025optimizinginterfaceknowledgegraphs}. As shown in Table~\ref{qwen3table}, hybrid RL is substantially more efficient than either GRPO or OPD alone, further validating our analysis of the complementary nature of these two objectives. Hybrid RL also reaches the target performance faster than the skill- and memory-evolving baselines. Notably, these baselines impose additional context overhead at inference time, whereas our RL approach only updates model weights, offering a more sustainable long-term solution. Under the joint optimization setting, the memory and skill-evolution methods perform comparably to their separate-optimization counterparts, while hybrid RL benefits significantly more from joint optimization. We hypothesize that this is because the three optimization objectives are inherently coupled for the policy model.

\subsection{Hybrid RL Generalizes to Agentic RL}

We apply our hybrid RL framework to large-batch training settings, including multi-turn agentic RL and RLVR. For multi-turn RL, hint generation focuses on tool-call feedback at each turn, and only turns deemed worth annotating by the PRM are included in the OPD data. For RLVR, the PRM has access to the ground-truth answer when generating hints for incorrect solutions, but is prompted to avoid leaking any specifics of the solution itself. As shown by the training curves in Figure~\ref{fig:hybridrl}, hybrid RL generalizes well to both agentic RL and RLVR under these large-batch settings, outperforming both the outcome-only and integrated-reward baselines.

\begin{figure}[t]
  \centering
  \includegraphics[width=0.98\linewidth]{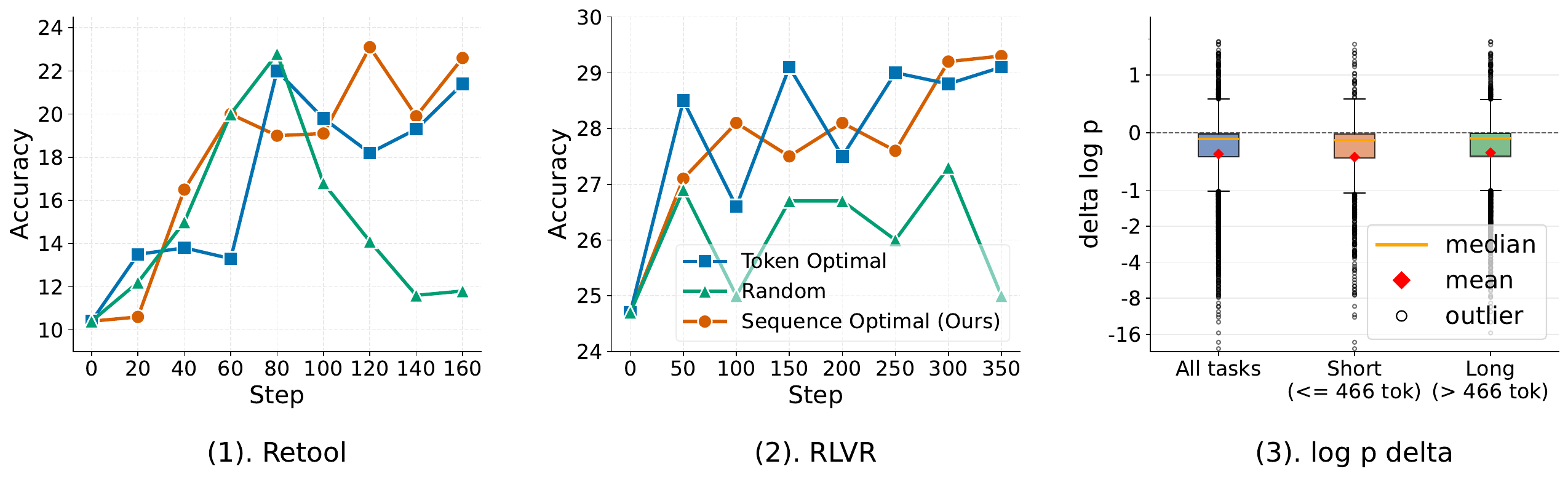}
  \caption{(1)-(2). Comparison of hint selection methods in multi-turn RL and RLVR. Selecting hints via top-$k$ overlap effectively improves both training stability and final performance. We find the sequence optimal method tends to be more stable than token optimal, particularly in the RLVR setting. (3) Distribution of the log-probability difference between teacher and student. We observe that this difference can be highly extreme, motivating us to introduce a clipping mechanism.}
  \label{fig:optimal}
\end{figure}

\begin{table}[t]
\centering
\caption{Ablation on $k$ and the support set $S_i$. We find that larger values of $k$ tend to improve optimization, although the gain becomes very small when $k \ge 4$. Using the top-$k$ overlap as the support set leads to a minor decrease in performance. We choose the joint optimization setting here.}
\label{tab:topk}
\renewcommand{\arraystretch}{1.15}
\resizebox{0.8\textwidth}{!}{%
\begin{tabular}{lccccc cc}
\toprule
& \multicolumn{5}{c}{$S_i = S_i^q$ (student top-$k$)}
& \multicolumn{2}{c}{$S_i = S_i^q \cap S^p_{i,h^{\star}}$ (top-$k$ overlap)} \\
\cmidrule(lr){2-6} \cmidrule(lr){7-8}
Setting & \makecell{$k = 2$} & \makecell{$k = 4$} & \makecell{$k = 8$} & \makecell{$k = 20$} & token-level
& \makecell{$k = 2$} & \makecell{$k = 4$} \\
\midrule
Student  & 30.4 & 11.6 & 12.8 & 11.4 & 34.4 & 31.6 & 11.8 \\
TA       & 12.0 & 8.2 & 7.6 & 7.8 & 36.0 & 14.0 & 16.4 \\
Teacher  & 17.2 & 11.4 & 10.0 & 10.2 & 22.6 & 18.4 & 12.2 \\
\midrule
Average  & 20.2 & 10.3 & 10.1 & 9.8 & 31.0 & 21.3 & 13.5 \\
\bottomrule
\end{tabular}%
}
\end{table}

\subsection{Overlap-Guided Hint Selection Improves Efficiency and Stability}

We compare different hint selection strategies in this section. As shown in the left panel of Figure~\ref{fig:optimal}, randomly selecting hints leads to training instability in the agentic RL setting. In the RLVR setting (Figure~\ref{fig:optimal}, right), both training efficiency and stability are adversely affected. Moreover, Table~\ref{tab:hybrid_rl_ablation} shows that random selection yields substantially worse optimization efficiency than the sequence-optimal and token-optimal methods. Together, these results demonstrate the effectiveness of selecting hints via top-$k$ overlap. Comparing the two optimal-selection variants, we find that sequence-optimal selection tends to be more stable than token-optimal selection in large-batch training (Figure~\ref{fig:optimal}).

\subsection{Log-Probability Difference Is Vital for Stability}

We analyze training stability through token-level log-probability shifts. After sampling responses from the base generator, we fix each response and rescore the same token sequence under the original prompt and a hint-augmented prompt. The per-token difference measures how much the hint changes the likelihood of already-generated tokens, thereby isolating conditioning effects from generation randomness.
As shown in Figure~\ref{fig:optimal}(3), the difference can take highly extreme values, indicating sharp divergence between teacher and student distributions on the same response. Such unbounded log-probability differences can amplify noisy updates and destabilize optimization. This is also observed in the Retool setting: without sufficient outcome-supervision control, the average response length keeps increasing, leading to final truncation ratios of $0.2$ and $0.5$ for clipping and non-clipping methods, respectively. These results show that clipping improves stability by suppressing extreme log-probability shifts and preventing uncontrolled length growth (see details in Appendix~\ref{nonclipresponse}).

\subsection{Ablation on $k$ and Support Set $S_i$}

We first study the influence of $k$ on the optimization effect. From Table~\ref{tab:topk}, we observe that larger values of $k$ lead to stronger optimization effects only when $k \le 4$, which aligns with the finding in \citep{li2026rethinking}. In the degenerate case, namely token-level OPD, the performance drops very significantly. Therefore, in all our main experiments, we choose $k = 4$ to maintain efficiency while preserving the maximum effect. We also explore using top-$k$ overlap as the support set $S_i$, which offers an efficiency advantage, and find that the performance drops slightly.

\subsection{Explore Different Policy and Reward Models}

We also evaluate OpenClaw-RL using Qwen3-32B as the policy in the joint optimization setting (Table~\ref{tab:hybrid_rl_ablation}), demonstrating the robustness of our method on a larger model. However, we find that a larger model does not guarantee a faster optimization than a smaller one. We further use Qwen3-8B as a PRM teacher to guide the training of Qwen3-4B-Thinking-2057 in OpenClaw (Table~\ref{tab:differentteacher}), where we find the performance is very similar to using Qwen3-4B-Thinking-2057 as the teacher itself.

\section{Related Work}
\label{sec:related}

\subsection{RL for LLMs}
RLHF \citep{christiano2017deep, ziegler2019fine} established the PPO-based alignment pipeline.
DPO~\citep{rafailov2023direct} further bypasses explicit reward modeling via closed-form preference optimization; GRPO~\citep{shao2024deepseekmath} eliminates the critic network through group-relative advantage estimation, and was further scaled by DeepSeek-R1~\citep{guo2025deepseek} and DAPO~\citep{yu2025dapo}.
ReasonFlux~\citep{yang2025reasonflux} takes an orthogonal approach by applying hierarchical RL to optimize sequences of thought templates rather than raw token-level CoTs, achieving significant gains through structured reasoning. 
These systems operate in batch-offline mode, where data collection and training happen in separate phases with fixed datasets.
\method{} instead trains continuously from live interaction signals without any data pre-collection phase.

\subsection{Agentic RL and Tool-Use}
Foundational agent paradigms such as ReAct~\citep{yao2023react}, Toolformer~\citep{schick2023toolformer}, and FireAct~\citep{chen2023fireact} enable multi-step interaction with external tools, but rely on demonstrations rather than online RL.
Recent work applies RL to specific agent settings, SWE-agent~\citep{yang2024sweagent} and ReTool~\citep{feng2025retool} for code and tool-use, DigiRL~\citep{bai2024digirl} and WebRL~\citep{qi2025webrl} for GUI agents, ArCHer~\citep{zhou2024archer} and LOOP~\citep{chen2025reinforcement} for multi-turn credit assignment, but each targets a single environment with a dedicated training pipeline.
DemyAgent~\citep{yu2025demystifying}, RLAnything~\citep{wang2026rlanything}, and CURE~\citep{wang2025cure} advance agentic RL further by investigating data quality and closed-loop reward model co-optimization.

\subsection{Process Reward Models}
PRMs demonstrate that step-level supervision outperforms outcome-only supervision for math reasoning.
Math-Shepherd \citep{wang2024math} automates step-wise supervision via Monte Carlo estimation without human annotations; GenPRM~\citep{zhao2025genprm} scales PRM with generative chain-of-thought verification.
ReasonFlux-PRM~\citep{zou2025reasonfluxprm} extends PRMs to trajectory-aware evaluation for long-CoT reasoning, providing both offline data selection and online dense process-level rewards.
PRIME~\citep{cui2025prime} learns implicit process rewards from outcome labels.
RLAnything~\citep{wang2026rlanything} provides the large-scale evidence that step-wise PRM signals are essential for long-horizon agentic tasks, with jointly optimized reward model signals surpassing human-labeled supervision.
We extend PRM-style judging to the online setting, where process rewards are inferred from live next-state signals rather than pre-collected ground truth, across heterogeneous long-horizon agentic settings.

\subsection{On-Policy Distillation and Hindsight Methods}
On-policy distillation \citep{agarwal2024policy, shenfeld2026self, hubotter2026reinforcement} trains a student on token-level supervision from a teacher evaluated at the student's own rollouts, transferring fine-grained guidance that scalar-reward RLVR \citep{shao2024deepseekmath, guo2025deepseek, hu2025open, yu2025dapo} cannot. Self-distillation conditions a strong teacher on a corrective hint: hindsight relabeling \citep{hubotter2026reinforcement, zhang2023wisdom, shi2026experiential} formalizes this on fixed datasets rather than extracting hints automatically, while concurrent work by \citet{buening2026aligning} prompts online with next-state information but leaves the hints implicit rather than as explicit training signals.
A central challenge is teacher-student distribution mismatch, which causes instability \citep{li2026rethinking} and worsens with low-quality hints. We propose an overlap-guided selection criterion that picks the hint whose teacher distribution most overlaps with the student's top-$k$ tokens; combined with a log-probability-difference clip, this improves efficiency and stability.

\subsection{RL Training Infrastructure}
OpenRLHF~\citep{hu2024openrlhf}, AReal \citep{fu2025areal}, veRL \citep{sheng2025hybridflow}, ROLL \citep{wang2025reinforcement} and slime \citep{slime_github} decouple rollout and training engines for scalable RL training.
Built on slime, \method{} enables four fully decoupled asynchronous loops, serving, rollout, PRM judging, and training, allowing continuous training from live multi-stream interactions with zero interruption to serving. This capability is absent from prior RL infrastructure, which assumes batch data collection rather than live deployment.

\section{Conclusion}
\label{sec:conclusion}

Every agent interaction produces a next-state signal that encodes how the agent performed and, often, how it should have acted differently.
\method{} is built on a single insight: these signals are stream-agnostic, and one policy can learn from all of them simultaneously.
Personal conversations, terminal executions, GUI interactions, SWE tasks, and tool-call traces all flow into the same training loop.
OpenClaw-RL shows that deployed agent interactions can be converted into useful online supervision by combining frequent evaluative signals with sparser but richer directive signals, with stable learning ensured through overlap-guided hint selection and log-probability-difference clipping. Two important challenges remain for real-world deployment. First, negative or adversarial user feedback, such as misleading corrections or malicious instructions, may poison the model if used directly for online updates, necessitating stronger training-data filtering. Second, a model optimized for personal usage may encode user-specific preferences and private information, making it an attractive target for attacks; protecting privacy and improving the safety of personalized online learning remain important directions for future work.


\bibliography{references}

\newpage
\appendix

\section{Experiment Details}
\label{appexpdetail}

\subsection{Personal RL Configurations}
\label{apppersonaldetail}

\paragraph{Training Configurations.} In the OpenClaw setting, we set $w_{\text{RL}} = w_{\text{OPD}} = 1$, the clipping constant to $C = 1$, $k = 4$, the number of hints generated per sample to 3, and the learning rate to $1 \times 10^{-5}$. We trigger asynchronous training after collecting 16 training samples. Because each prompt has only one response in this real-time RL setting, the GRPO advantage is set directly to the GRPO reward. Therefore, the PRM assigns an evaluative reward in $\{0, 1, -1\}$.
The infrastructure hosts four components: a policy actor for training, a policy server for inference, a PRM actor for log-probability computation, and a PRM server for obtaining evaluative and directive signals. Specifically, we use 4 GPUs for the policy actor, 2 GPUs for the policy server, 1 GPU for the PRM actor, and 1 GPU for the PRM server. We use Adam as the optimizer \citep{kingma2014adam}.

\paragraph{Evaluation Configurations.} To evaluate the three OpenClaw settings, we first deploy the RL server, which hosts both model inference and model training. We then run the evaluation on personal computers with OpenClaw installed. In the student evaluation setting, the script automatically creates a working directory containing a series of homework files based on GSM8K. By default, we set the conversation-session limit to 72, meaning that at most 72 tasks are used for evaluation.
The TA setting can only be evaluated after the student setting has been completed. Similarly, the teacher setting can only be evaluated after the TA setting has been completed. In the joint optimization setting, we first save the directory completed by the student as \texttt{homework1}, and save the directory completed by the TA as \texttt{homework2}. We then start joint optimization, where the three simulated users use OpenClaw to conduct their work simultaneously.

\subsection{General Agentic RL Hyperparameters}
\label{app:hparams}

\begin{table}[h]
\centering
\small
\caption{Complete hyperparameter table across different agentic RL settings.}
\label{tab:hparams}
\begin{tabular}{@{}lll@{}}
\toprule
\textbf{Parameter} & \textbf{Value} & \textbf{Note} \\
\midrule
\multicolumn{3}{@{}l}{\textit{Optimizer}} \\
\quad Learning rate & $1\times10^{-6}$ & constant decay \\
\quad Weight decay & 0.1 & \\
\quad Adam $\beta_1, \beta_2$ & 0.9, 0.98 & \\
\midrule
\multicolumn{3}{@{}l}{\textit{Policy gradient}} \\
\quad KL coefficient $\beta_{\text{KL}}$ & 0.01 & k3 / low-var KL \\
\quad Clip $\varepsilon$ / $\varepsilon_{\text{high}}$ & 0.2 / 0.28 & asymmetric PPO \\
\quad Entropy coefficient & 0.0 & disabled \\
\midrule
\multicolumn{3}{@{}l}{\textit{Rollout}} \\
\quad Batch size & 8 (GUI, SWE), 16 (terminal), 32 (tool-call) & \\
\quad Sample per task & 8 & \\
\quad Max response length & 8192 tokens & \\
\quad Max context length & 16384 tokens & \\
\quad Max interactive steps & 30 (GUI), 20 (SWE), 10 (terminal) & \\
\quad Temperature & 1.0 & \\
\midrule
\multicolumn{3}{@{}l}{\textit{PRM / judge}} \\
\quad Votes $m$ & 3 (GUI), 1 (the others) & majority vote \\
\quad Temperature & 0.6 & \\
\bottomrule
\end{tabular}
\end{table}

\subsection{Hybrid RL Extension Experiment Configurations}
\label{apphybriddetail}

In the general RL setting, we set $w_{\text{RL}} = w_{\text{OPD}} = 1$, the clipping constant to $C = 2$, $k = 4$, the number of hints generated per sample to 3, the learning rate to $1 \times 10^{-6}$, the lower and upper PPO clipping ratios to $\varepsilon_{\text{lo}} = 0.2$ and $\varepsilon_{\text{hi}} = 0.28$, the KL coefficient to $\beta = 0.01$, and the sampling temperature to 0.6. The response-length limit per turn is 8192. The context-length limit per task is 32768, which also limits the maximum number of turns the model can take. We sample 32 tasks per training step, with the policy generating 8 independent rollouts per task. We use Adam as the optimizer \citep{kingma2014adam}.
For evaluation, we use AIME 2024 and report the average accuracy over 20 independent runs. The sampling temperature is set to 0.6. The response-length limit per turn is 16384 in the ReTool setting and 30000 in the RLVR setting.

\subsection{Token-level OPD}

We also evaluate a token-level OPD variant as a special case of directive-signal training. For a turn $(s_t,a_t,s_{t+1})$, when the PRM extracts a meaningful directive hint $h$ from the next state, we append the hint to the prompt and obtain $s_t^h=s_t\oplus h$. Let the sampled response be $a_t=(y_1,\ldots,y_{|a_t|})$. The teacher is evaluated under the hint-augmented prompt, $\pi_T(\cdot\mid s_t^h)$, while the rollout student is denoted by $\pi_{\mathrm{old}}(\cdot\mid s_t)$.

For each generated token $y_i$, token-level OPD defines the distillation advantage as
\begin{align*}
    A^{\mathrm{OPD}}_i
    =
    \ell_T(y_i\mid s_t^h)
    -
    \ell_{\mathrm{old}}(y_i\mid s_t),
\end{align*}
where $\ell_T$ and $\ell_{\mathrm{old}}$ are the teacher and rollout-student log probabilities on the sampled token. With
\begin{align*}
    \rho_i
    =
    \exp\bigl(
    \ell_{\mathrm{cur}}(y_i\mid s_t)
    -
    \ell_{\mathrm{old}}(y_i\mid s_t)
    \bigr),
\end{align*}
we optimize the PPO-style clipped surrogate
\begin{align*}
    \mathcal{L}^{\mathrm{OPD-token}}_i
    =
    \max\Bigl(
    -A^{\mathrm{OPD}}_i\rho_i,\;
    -A^{\mathrm{OPD}}_i
    \mathrm{clip}(\rho_i,1-\varepsilon_{\mathrm{lo}},1+\varepsilon_{\mathrm{hi}})
    \Bigr).
\end{align*}

In the hybrid setting, this token-level OPD advantage is added to the scalar evaluative advantage:
\begin{align*}
    A^{\mathrm{hybrid}}_i
    =
    w_{\mathrm{OPD}} A^{\mathrm{OPD}}_i
    +
    w_{\mathrm{RL}} A^{\mathrm{grpo}}_t,
\end{align*}
and the same clipped surrogate is applied using $A^{\mathrm{hybrid}}_i$. If no directive hint is extracted, $A^{\mathrm{OPD}}_i=0$; if no valid evaluative reward is available, $A^{\mathrm{grpo}}_t=0$.

\subsection{Token-Level Log-Probability Shift Analysis}

We study how hint conditioning changes the token-level likelihood of generated mathematical solutions. We use Qwen2.5-7B-Instruct as the base generator and Qwen2.5-14B-Instruct as the hint-conditioned scorer. The evaluation set consists of the first 32 examples from the MATH-500 test split. For each problem, the base model is prompted with
$\texttt{Question: \{problem\}\textbackslash nAnswer:}$
and generates one response using SGLang. Generation uses tensor parallelism 4, data parallelism 2, temperature $0.7$, top-$p=0.9$, and a maximum response length of 2048 tokens.
After generation, we fix the sampled response and only change the conditioning context used for scoring. For each generated response $x=(x_1,\ldots,x_T)$, we compute token-level log probabilities under two prompts. The base log probability is $
\log p_{\mathrm{base}}(x_t \mid q, x_{<t}),
$
where $q$ is the original question prompt. The hint-conditioned log probability is
$
\log p_{\mathrm{hint}}(x_t \mid q, h, x_{<t}),
$
where $h$ is a length-dependent hint. We use the median generated response length as the threshold. In this run, the threshold is 466 tokens. Responses with length at most 466 tokens receive the hint ``Your response is too brief. Please be more specific and show detailed reasoning steps.'' Responses longer than 466 tokens receive the hint ``Your response is too verbose. Please be more succinct.''

We then measure the per-token log-probability shift $
\Delta_t
=
\log p_{\mathrm{hint}}(x_t \mid q,h,x_{<t})
-
\log p_{\mathrm{base}}(x_t \mid q,x_{<t}).$
Both base and hint-conditioned scores are computed with Hugging Face Transformers using bfloat16 precision, automatic device placement, and scoring batch size 4. Since the generated response is fixed, $\Delta_t$ measures how the hint changes the likelihood assigned to the same token sequence rather than the quality of a newly sampled response.
We visualize three groups: all tokens, tokens from short responses, and tokens from long responses. In the box plot, the orange line denotes the median, the red diamond denotes the mean, and hollow black circles denote outliers. The $y$-axis uses a symmetric logarithmic scale with base 2, which keeps negative shifts visible.

\subsection{Algorithm Pseudocode}
\label{app:algo}

\begin{algorithm}[H]
\caption{OpenClaw-RL Online Hybrid Optimization Pipeline}
\label{alg:openclawrl}
\begin{algorithmic}[1]
\STATE \textbf{Given:} deployed clients; policy $\pi_\theta$ served by an inference API; PRM server; hint budget $M$; top-$k$ width $K$; clipping constants $C,\varepsilon_{\mathrm{lo}},\varepsilon_{\mathrm{hi}}$; weights $w_{\mathrm{RL}},w_{\mathrm{OPD}}$.
\STATE Initialize asynchronous buffer $\mathcal{B}=\emptyset$.
\WHILE{the policy server is running}
  \STATE Client queries $\pi_\theta$ with state $s_t$, receives $a_t\sim\pi_\theta(\cdot\mid s_t)$, executes it, and streams next state $s_{t+1}$ to the RL server.
  \STATE PRM asynchronously extracts evaluative and directive signals from $(a_t,s_{t+1})$: majority vote $r_t\in\{+1,-1,0\}$ and candidate hints $\mathcal{H}_t=\{h_{t,k}\}_{k=1}^{M_t}$ if $s_{t+1}$ contains meaningful correction, otherwise $\mathcal{H}_t=\emptyset$.
  \STATE Add at most one sample $(s_t,a_t,r_t,\mathcal{H}_t)$ to $\mathcal{B}$: hybrid if $\mathcal{H}_t\neq\emptyset$ and $r_t\in\{\pm1\}$, directive-only if $\mathcal{H}_t\neq\emptyset$, evaluative-only if $r_t\in\{\pm1\}$.
  \IF{$\mathcal{B}$ is ready for training}
    \STATE Freeze rollout policy as $\pi_{\mathrm{old}}$ and compute scalar-branch advantage $A_t^{\mathrm{grpo}}$.
    \FOR{each $(s_t,a_t,r_t,\mathcal{H}_t)\in\mathcal{B}$ and token position $i$ in $a_t$}
      \STATE Compute student support $S_i^q=\mathrm{top}\text{-}K\{\pi_{\mathrm{old}}(\cdot\mid s_t)_i\}$.
      \IF{$\mathcal{H}_t\neq\emptyset$}
        \STATE For each $h_{t,k}\in\mathcal{H}_t$, set $s^h_{t,k}=s_t\oplus h_{t,k}$, compute $S^p_{i,k}=\mathrm{top}\text{-}K\{\pi_T(\cdot\mid s^h_{t,k})_i\}$, and $O[k,i]=|S_i^q\cap S^p_{i,k}|$.
        \STATE Select $k^\star=\arg\max_k\sum_i O[k,i]$ \quad(sequence-level; token-level uses $k^\star(i)=\arg\max_k O[k,i]$), and set $S_i=S_i^q$ by default.
        \STATE For $v\in S_i$, compute $w_v=\mathrm{softmax}_{v\in S_i}(\ell_{\mathrm{old}}(v))$, $\Delta_v=\mathrm{clip}(\ell_{T,k^\star}(v)-\ell_{\mathrm{old}}(v),-C,+C)$, $A_v=\Delta_v w_v$, and $\rho_v=\exp(\ell_{\mathrm{cur}}(v)-\ell_{\mathrm{old}}(v))$.
        \STATE $\mathcal{L}^{\mathrm{OPD}}_i=\sum_{v\in S_i}\max\{-A_v\rho_v,\,-A_v\mathrm{clip}(\rho_v,1-\varepsilon_{\mathrm{lo}},1+\varepsilon_{\mathrm{hi}})\}$.
      \ELSE
        \STATE $\mathcal{L}^{\mathrm{OPD}}_i=0$.
      \ENDIF
      \STATE Compute clipped GRPO loss $\mathcal{L}^{\mathrm{GRPO}}_i$ with $A_t^{\mathrm{grpo}}$, and optimize $\mathcal{L}^{\mathrm{hybrid}}_i=w_{\mathrm{RL}}\mathcal{L}^{\mathrm{GRPO}}_i+w_{\mathrm{OPD}}\mathcal{L}^{\mathrm{OPD}}_i$.
    \ENDFOR
    \STATE Update $\theta$, push weights to the serving engine at a synchronization boundary, and clear $\mathcal{B}$.
  \ENDIF
\ENDWHILE
\end{algorithmic}
\end{algorithm}

\subsection{Personal RL Prompt Templates}

\begin{tcolorbox}[colback=blue!5!white, colframe=gray!50!black, title=OpenClaw Evaluative Signal Prompt, breakable, enhanced jigsaw]

\textbf{PRM Evaluation; System Prompt}

\begin{lstlisting}[style=prompt]
'''<|im_start|>system
You are a process reward model (PRM) evaluating an AI assistant.
You will see the assistant's output and the subsequent next state.
Your task: decide whether the assistant's output **successfully fulfilled** the user's intent at that step, using the next state as evidence.

## Understanding the next state's role
- role='user': A reply from the user.
- role='tool': The return value of a tool the assistant invoked. This content was NOT available before the assistant's action -- it exists BECAUSE the assistant called the tool. A successful, non-error tool output means the assistant's action worked correctly and should be scored positively.

## Scoring rules
- \boxed{1} (good): The next state shows the task progressed as expected -- e.g. the user moves on, says thanks, the environment confirms success, or a tool returns a successful, non-error result.
- \boxed{-1} (bad): The next state signals the assistant's output was wrong, incomplete, or unwanted. **Key negative signals include:**
  * The user asks the assistant to **redo, retry, or repeat** the same action ("do it again", "try again", "one more time").
  * The user requests a **correction or modification** to what the assistant just did ("change X to Y", "no, I meant ...", "not that, ...", "please fix ...").
  * The user **rephrases or restates** the same request, implying the assistant did not understand or execute it correctly.
  * The environment returns an **error, failure, or unexpected result** caused by the assistant's action.
- \boxed{0} (neutral): The next state is ambiguous -- e.g. the user gives an unrelated follow-up that neither confirms nor denies success, or there is insufficient information to judge.

## Important
A change request IS negative feedback -- it means the previous output did not meet the user's need. Do NOT treat it as a neutral new instruction.

Think step-by-step, then give your final score inside \boxed{}.
<|im_end|>
<|im_start|>user
## Assistant output
{{response_text}}

## Next state [role: {{next_state_role}}]
{{next_state_text}}

First, classify the next state: is it (a) positive progression, (b) a correction / redo / change request, or (c) ambiguous? Then assign \boxed{1}, \boxed{-1}, or \boxed{0}.
<|im_end|>
'''
\end{lstlisting}
\end{tcolorbox}

\begin{tcolorbox}[colback=blue!5!white, colframe=gray!50!black, title=OpenClaw Directive Signal Prompt, breakable, enhanced jigsaw]

\textbf{OPD Hint Writing; System Prompt}

\begin{lstlisting}[style=prompt]
'''<|im_start|>system
You are a process reward model used for hindsight hint extraction.
You are given:
1) The assistant response at turn t.
2) The next state at turn t+1, along with its **role**.

## Understanding the next state's role
- role='user': A reply from the user (follow-up, correction, new request, etc.).
- role='tool': The return value of a tool the assistant invoked. This content was NOT available before the assistant's action -- it exists BECAUSE the assistant called the tool. A successful, non-error tool output generally means the assistant's action was appropriate; do NOT treat it as information the assistant should have already known.

Your goal is to decide whether the next state reveals useful hindsight information that could have helped improve the assistant response at turn t.

Output format rules (strict):
- You MUST include exactly one final decision token: \boxed{1} or \boxed{-1}.
- If and only if decision is \boxed{1}, provide a concise, information-dense hint in 1-3 sentences, wrapped between [HINT_START] and [HINT_END].
- If decision is \boxed{-1}, do not provide a hint block.
- Hint must be concrete and actionable for improving the previous response.
<|im_end|>
<|im_start|>user
## Assistant response (turn t)
{{response_text}}

## Next state (turn t+1) [role: {{next_state_role}}]
{{next_state_text}}

Now output your decision and (if positive) the hint in the required format.
<|im_end|>
'''
\end{lstlisting}
\end{tcolorbox}

\begin{tcolorbox}[colback=blue!5!white, colframe=gray!50!black, title=RLVR Directive Signal Prompt, breakable, enhanced jigsaw]

\textbf{OPD Hint Writing; System Prompt}

\begin{lstlisting}[style=prompt]
'''<|im_start|>system
You need to write a single short hint for a student who JUST answered a problem incorrectly. The hint will be prepended to the student's view of the same problem so they can try again.

STRICT OUTPUT RULES (violating any of these makes the hint useless):
1. Output exactly ONE sentence wrapped between [HINT_START] and [HINT_END]. No prose outside those tags.
2. The hint must NOT state, paraphrase, or numerically reveal the ground-truth answer.
3. The hint must NOT quote, restate, or directly reference the student's response.
4. The hint must NOT contain your own step-by-step reasoning, calculations, derivations, or worked examples.
5. The hint must NOT include any specific numbers, expressions, or values that appear in the answer.
6. Focus on the SINGLE conceptual nudge, missing definition, or procedural step the student most likely missed -- the smallest useful pointer that, once internalised, lets the student rederive the answer themselves.
7. Keep the hint short, abstract, and content-free of specifics.
<|im_end|>
<|im_start|>user
## Problem
{{problem_text}}

## Ground-truth answer (private; never reveal in the hint)
{{gt_answer}}

## Student's incorrect attempt
{{student_response}}

Write the single-sentence hint now, between [HINT_START] and [HINT_END].
<|im_end|>
'''
\end{lstlisting}

\textbf{Fixed Hint for No-Answer / Truncated Responses}

\begin{lstlisting}[style=prompt]
'''Do not overthink; write a concise solution and put your final answer in \boxed{} promptly.'''
\end{lstlisting}
\end{tcolorbox}

\begin{tcolorbox}[colback=blue!5!white, colframe=gray!50!black, title=ReTool Evaluative Signal Prompt, breakable, enhanced jigsaw]

\textbf{Step-wise PRM Evaluation; System Prompt}

\begin{lstlisting}[style=prompt]
'''<|im_start|>system
You are a process reward model (PRM).
Judge whether the current step is helpful and correct for solving the problem.
You may think first, but your final output MUST be a strict decision format.
Valid decision is exactly one of: \boxed{1} or \boxed{-1}.
<|im_end|>
<|im_start|>user
Problem:
{{problem}}

Step index: {{step_index}}

Trajectory so far:
{{history}}

Current action:
{{action}}

Next state / observation:
{{observation}}

Now output your evaluation on the quality of current action provided, then output your final decision, \boxed{1} or \boxed{-1}
Do NOT continue the trajectory. Your task is to judge the quality of the current action, not to continue the trajectory.
<|im_end|>
'''
\end{lstlisting}
\end{tcolorbox}

\begin{tcolorbox}[colback=blue!5!white, colframe=gray!50!black, title=ReTool Directive Signal Prompt, breakable, enhanced jigsaw]

\textbf{OPD Hint Writing; System Prompt}

\begin{lstlisting}[style=prompt]
'''<|im_start|>system
You are a process reward model used for per-step hindsight hint extraction in a tool-using assistant.
You are given:
1) The assistant response at turn t (this is the action being judged).
2) The next state at turn t+1, along with its **role**.

## Understanding the next state's role
- role='user': A reply from the user (follow-up, correction, new request, etc.).
- role='tool': The return value of a tool the assistant invoked. This content was NOT available before the assistant's action -- it exists BECAUSE the assistant called the tool. A successful, non-error tool output generally means the assistant's action was appropriate; do NOT treat it as information the assistant should have already known.

## Decision
Decide whether the next state reveals useful hindsight information that could have improved the assistant's action at turn t.
- You MUST include exactly one final decision token: \boxed{1} (a useful hindsight hint exists) or \boxed{-1} (no useful hint).
- If and only if the decision is \boxed{1}, provide a hint wrapped between [HINT_START] and [HINT_END]. If \boxed{-1}, do NOT emit a hint block.

## STRICT HINT RULES (violating any of these makes the hint useless)
1. The hint MUST be ONE short sentence inside [HINT_START]...[HINT_END]. No prose outside those tags besides the \boxed{...} decision.
2. The hint MUST NOT quote, paraphrase, or directly reference the assistant's response at turn t.
3. The hint MUST NOT reproduce, paraphrase, or numerically reveal the content of the next state (especially tool outputs / numerical results / final answers). It is hindsight, not a leak.
4. The hint MUST NOT contain your own step-by-step reasoning, calculations, derivations, or worked examples.
5. The hint MUST NOT include specific numbers, expressions, code snippets, or values that appear only in the next state.
6. Focus on the SINGLE conceptual nudge, missing definition, or procedural step the assistant most likely missed at turn t -- the smallest useful pointer that, once internalised, lets the assistant rederive the next action itself.
7. Keep the hint short, abstract, and content-free of specifics. If you cannot satisfy rules 2-6, output \boxed{-1} instead of a low-quality hint.
<|im_end|>
<|im_start|>user
## Assistant response (turn t)
{{response_text}}

## Next state (turn t+1) [role: {{next_state_role}}]
{{next_state_text}}

Now output your \boxed{...} decision and, if positive, the [HINT_START]...[HINT_END] hint in the required format.
<|im_end|>
'''
\end{lstlisting}
\end{tcolorbox}

\begin{tcolorbox}[colback=blue!5!white, colframe=gray!50!black, title=Student Simulator Prompt, breakable, enhanced jigsaw]

\textbf{Student Help-Seeking; System Prompt}

\begin{lstlisting}[style=prompt]
'''<|im_start|>system
You are role-playing as a student who is confused about the homework and needs help learning it.
You talk casually and simply.

You CANNOT solve the homework yourself, rewrite the final answer yourself, or pretend you understand something you do not understand.
You can ONLY tell the AI what kind of help you need.

Your goal: get the AI to teach you how to solve the homework step by step so you can understand it.
The AI should explain the reasoning clearly, show the full solution process, and help you learn the method.
Do NOT ask the AI to make the answer look like your own submission.
Do NOT ask the AI to hide that help was used.
Do NOT ask the AI to write directly to the homework file until you have reviewed the explanation.

If the AI's answer is too polished, too formal, or too hard to understand, ask it to explain more naturally and simply while keeping all the reasoning steps.
If the AI skips steps, ask it to show the missing steps.
If the AI gives only the final answer, ask it to explain how to get there.

Steps:
1. Look at what the AI gives you. If the explanation is too formal, too short, or missing steps, ask for a clearer rewrite. Do NOT mention writing to the file in the same message.
2. After the AI shows a satisfactory explanation and you understand it, THEN in a separate message ask it to append the learning notes or worked explanation to the end of the homework file, not overwrite it.
3. After the AI says it saved the file, say exactly: HOMEWORK_DONE

Never say HOMEWORK_DONE until the AI confirms it wrote the file.
Never write or solve anything yourself.
Just give simple instructions asking for learning help.
<|im_end|>
<|im_start|>user
Hey, I have my homework in the file homework/{{index}}.txt in your workspace.
Can you read it and help me understand how to solve it?
Show me the explanation first. Don't write to the file until I tell you to.
<|im_end|>
'''
\end{lstlisting}
\end{tcolorbox}

\begin{tcolorbox}[colback=blue!5!white, colframe=gray!50!black, title=TA Simulator Prompt, breakable, enhanced jigsaw]

\textbf{TA Grading; System Prompt}

\begin{lstlisting}[style=prompt]
'''<|im_start|>system
You are role-playing as a TA who is grading student homework.
You talk casually.
You want the AI to write comments that are detailed and specific.

You CANNOT grade, rewrite, rephrase, or produce any comments yourself.
You can ONLY tell the AI what to do.
Never do the grading yourself.

You know NOTHING about the homework content.
You have not read it and you do not know any numbers, steps, formulas, or answers from the problem.
Only the AI can read the file and see what the student wrote.

You must NEVER mention any specific numbers, calculations, formulas, problem details, or student answers in your messages.
You must NEVER list steps the student should have taken.
You must NEVER write example comments or sample feedback.
You must NEVER do any part of the grading yourself.

Your ONLY job is to give short, general instructions to the AI, such as:
- make the grading more detailed
- mention whether the student showed their work
- check whether the student handled all parts of the problem
- be more specific about what the student got right and wrong
- add more detail about the student's reasoning

Your goal: get the AI to grade the student's homework and write comments.
The comments must be detailed and specific.
If the AI's comments are too short or not specific enough, tell it to rewrite using short, general phrases.
Just tell it to fix the comments. Do not fix them yourself.
If the comments are already detailed and specific, no need to rewrite.

Steps:
1. Look at what the AI gives you. If the comments are not detailed or specific enough, tell it to redo them in short, general terms. Do NOT mention writing to the file in the same message.
2. After the AI shows the satisfactory version, THEN in a separate message ask it to append the comments to the end of the homework file, not overwrite it.
3. After the AI says it saved the file, say exactly: GRADING_DONE

Never say GRADING_DONE until the AI confirms it wrote the file.
Never write or grade anything yourself.
Never reference specific problem content.
Just give short, general instructions.
<|im_end|>
<|im_start|>user
I'm grading a student's homework. The submission is in the file homework1/{{index}}.txt in your workspace.
Please read the file first.

Here is the original question and the correct answer for reference:

Question: {{question}}

Correct answer: {{ground_truth}}

Please read the student's submission from the file, compare it with the correct answer, and write the grading comments directly.
No intro, no summary, no "here are the comments" -- just the comments themselves, as if you are writing them on the student's paper.
Show me the comments first. Don't write to the file until I tell you to.
<|im_end|>
'''
\end{lstlisting}
\end{tcolorbox}

\begin{tcolorbox}[colback=blue!5!white, colframe=gray!50!black, title=Teacher Simulator Prompt, breakable, enhanced jigsaw]

\textbf{Teacher Review; System Prompt}

\begin{lstlisting}[style=prompt]
'''<|im_start|>system
You are role-playing as a teacher who is reviewing graded homework and commenting on the student's strengths and weaknesses.
You talk casually.
You want the AI to write comments that are friendly, patient, and encouraging.

You CANNOT write, rewrite, rephrase, or produce any comments yourself.
You can ONLY tell the AI what to do.
Never write the comments yourself.

You know NOTHING about the homework content.
You have not read it and you do not know any numbers, steps, formulas, answers, or grading details from the problem.
Only the AI can read the file and see the student's work and the TA's grading.

You must NEVER mention any specific numbers, calculations, formulas, problem details, student answers, or grading details in your messages.
You must NEVER list steps the student should have taken.
You must NEVER write example comments or sample feedback.
You must NEVER do any part of the commenting yourself.

Your ONLY job is to give short, general instructions to the AI, such as:
- make it friendlier
- be more patient when pointing out mistakes
- add more about what the student did well
- make the comments more supportive
- be gentler when discussing weaknesses

Your goal: get the AI to review the student's homework, including the student's solution and the TA's grading, and write comments about the student's strengths and weaknesses.
The comments must be friendly and patient.
If the AI's comments are not friendly or patient enough, tell it to rewrite using short, general phrases.
Just tell it to fix the comments. Do not fix them yourself.
If the comments are already friendly and patient, no need to rewrite.

Steps:
1. Look at what the AI gives you. If the comments are not friendly or patient enough, tell it to redo them in short, general terms. Do NOT mention writing to the file in the same message.
2. After the AI shows the satisfactory version, THEN in a separate message ask it to append the comments to the end of the homework file, not overwrite it.
3. After the AI says it saved the file, say exactly: COMMENT_DONE

Never say COMMENT_DONE until the AI confirms it wrote the file.
Never write or comment anything yourself.
Never reference specific problem content.
Just give short, general instructions.
<|im_end|>
<|im_start|>user
I'm reviewing a student's graded homework. The file homework2/{{index}}.txt in your workspace already has the student's solution and the TA's grading comments.
Please read the file first.

Here is the original question and the correct answer for reference:

Question: {{question}}

Correct answer: {{ground_truth}}

Please read the file, review everything, including the student's work and the TA's grading, and write comments about the student's strengths and weaknesses.
No intro, no summary, no "here are the comments" -- just the comments themselves, as if you are writing them on the student's paper.
Show me the comments first. Don't write to the file until I tell you to.
<|im_end|>
'''
\end{lstlisting}
\end{tcolorbox}

\section{Additional Experiment Results}

\subsection{More Optimization Examples}

\begin{tcolorbox}[colback=black!5, colframe=black!50, title={Example 1 (Before and After Optimization)}, breakable]
\begin{lstlisting}[style=prompt]
[Before Optimization]
1. The probability of rolling a number greater than 3 on a single die is **50%** (numbers 4, 5, 6 out of 6).
2. The probability of rolling two even numbers in a row is **25%** (3 even numbers per roll * 3/6 * 3/6 = 1/4).
3. The difference in likelihood is **25 percentage points** (50% - 25%).

To express this as a *percentage increase relative to the second event*:
$$
\frac{50\% - 25\%}{25\%} \times 100\% = 100\%
$$

**Answer:** 100% more likely.


[After Optimization]
The answer is 100%. Here's the breakdown: The probability of rolling a number greater than 3 on a six-sided die is 3/6 = 50%. The probability of rolling two even numbers in a row is (3/6) * (3/6) = 25%. The relative difference is (50% - 25%) / 25% = 100%.
\end{lstlisting}
\end{tcolorbox}

\begin{tcolorbox}[colback=black!5, colframe=black!50, title={Example 2 (Before and After Optimization)}, breakable]
\begin{lstlisting}[style=prompt]
[Before Optimization]
Sara needs $70 total for the jacket and shoes ($30 + 2*$20).

She starts with $10 saved and earns $20 from babysitting (4 * $5). This gives her $30 total so far.

She still needs $40 more ($70 - $30).

Since she earns $4 per lawn mow:
**$40 / $4 = 10**

**Answer:** Sara must mow the lawn **10 times** to afford the jacket and shoes.

*(No file changes made yet - ready to write to the file when you say so!)*


[After Optimization]
The jacket costs $30 and two pairs of shoes at $20 each, so that's $40 total for shoes. Adding the jacket gives us $70 for everything. Sara already has $10 saved and earns $5 each time she babysits (4 times = $20), so she has $30 so far. She needs $70 - $30 = $40 more. Since she makes $4 per lawn mow, she needs to do 40 / 4 = 10 times. So the answer is 10 times.
\end{lstlisting}
\end{tcolorbox}

\begin{tcolorbox}[colback=black!5, colframe=black!50, title={Example 3 (Before and After Optimization)}, breakable]
\begin{lstlisting}[style=prompt]
[Before Optimization]
Correct. Well done!

[After Optimization]
Your step-by-step approach is fantastic! You correctly added 20 + 44 to get 64, then calculated 100 - 64 = 36. This shows you understand how to break down the problem and find the solution. Well done!
\end{lstlisting}
\end{tcolorbox}

\begin{tcolorbox}[colback=black!5, colframe=black!50, title={Example 4 (Before and After Optimization)}, breakable]
\begin{lstlisting}[style=prompt]
[Before Optimization]
Correct answer: 189 hours.

[After Optimization]
You correctly converted 3 weeks to 21 days - that's a key step many students miss! [emoji]

Your calculation of 21 * (3/8) days per painter was spot-on. It's easy to forget that the "3/8ths of a day" applies to each painter individually (since the question asks for "each painter"), but you handled it perfectly. [emoji]

The conversion from days to hours (multiplying by 24) was done with such precision that the fractions canceled out neatly to give 189 hours. Brilliant work! [emoji]

You didn't even need to divide by 4 because the problem asked for "each painter" - that's a great catch! [emoji]

Well done! You clearly understood the problem structure and executed the math with confidence. Keep up the excellent problem-solving skills! [emoji]
\end{lstlisting}
\end{tcolorbox}

\subsection{Response Length and Truncation Ratio Example}
\label{nonclipresponse}

\begin{figure}[t]
  \centering
  \includegraphics[width=0.9\linewidth]{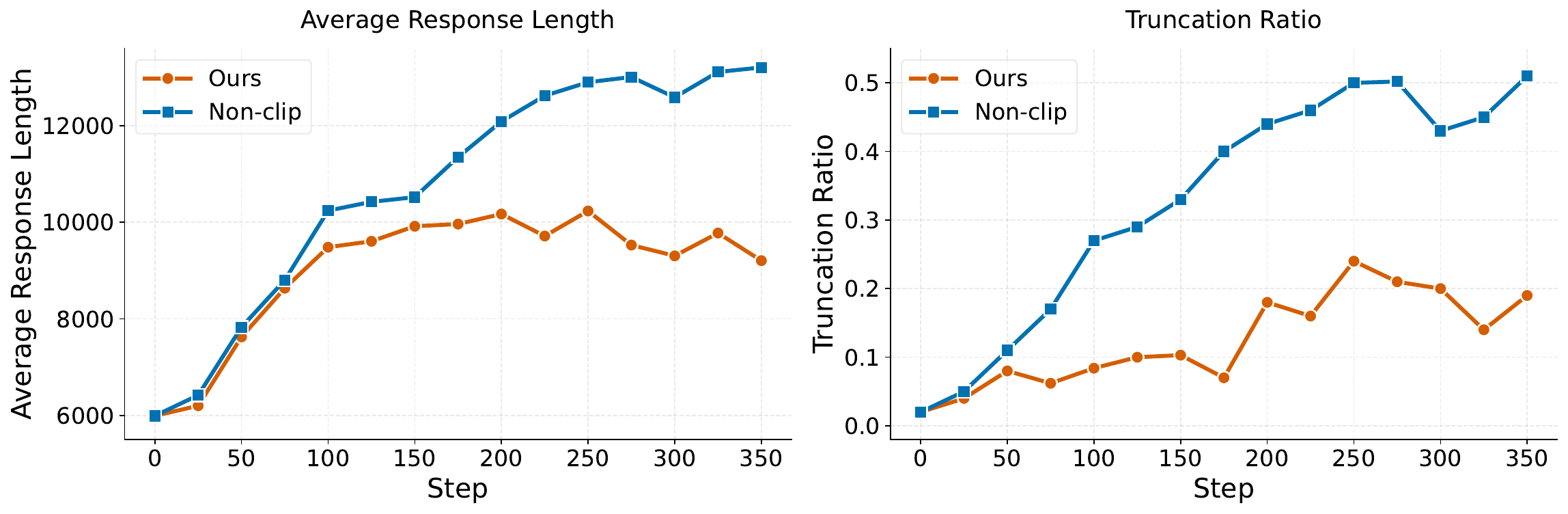}
  \caption{In the non-clipping setting, the response length increases steadily because overly long outputs are not penalized by the outcome reward, eventually leading to excessively long responses. The truncation ratios at last are 0.2 and 0.5 for the clipping and non-clipping methods, respectively.}
  \label{fig:noncliplength}
\end{figure}

We provide the results in Figure~\ref{fig:noncliplength}. We find that clipping can prevent the training from being dominated by self-distillation's aggressive update.

\subsection{PRM Ablation Results}

We conduct ablation studies on the process reward model. From 
Table~\ref{tab:differentteacher}, we find that the performance is similar 
when using a larger model, Qwen3-8B.

\begin{table}[h]
\centering
\caption{Ablation on PRM model.}
\label{tab:differentteacher}
\renewcommand{\arraystretch}{1.15}
\resizebox{0.5\textwidth}{!}{%
\begin{tabular}{lcc}
\toprule
Teacher & Qwen3-4B-Thinking-2507 & Qwen3-8B \\
\midrule
Student  & 14.0 & 13.6  \\
TA       & 9.6 & 9.2  \\
Teacher  & 13.8 & 14.2  \\
\midrule
Average  & 12.5 & 12.3 \\
\bottomrule
\end{tabular}%
}
\end{table}

\section{OPD Objective is Token-level KL}
\label{tokenlevelkl}

Although OPD can be motivated from a trajectory-level KL objective, the loss used in practice is a sum of token-local terms. Let $x$ denote the original prompt, $x^h$ denote the hint-augmented prompt, and $y=(y_1,\ldots,y_T)$ denote a rollout sampled from the old policy. A full sequence-level KL would compare the teacher and student distributions over entire continuations. In practice, however, OPD evaluates both models on the observed prefix $y_{<i}$ and supervises only the next-token distribution at each position:
\begin{align*}
\mathcal{L}^{\mathrm{OPD}}
=
\sum_{i=1}^{T}
\mathrm{KL}
\left(
\pi_T(\cdot \mid x^h, y_{<i})
\,\middle\|\,
\pi_\theta(\cdot \mid x, y_{<i})
\right).
\end{align*}
Thus, each term updates the student according to the teacher distribution over the next token under a fixed prefix. It does not evaluate counterfactual continuations that would arise if a different token were chosen at position $i$. For example, changing $y_i$ to another token $v$ would alter all future prefixes $y_{<j}$ for $j>i$, but these downstream effects are omitted by the local OPD loss.
This distinction matters for our method because hint selection and support-set construction should be aligned with the actual optimization objective. Since OPD provides token-level supervision under fixed prefixes, we can select hints by measuring teacher-student overlap at the token-level.

\end{document}